\documentclass[lettersize,journal]{IEEEtran} 

\usepackage{multirow}
\usepackage{makecell}
\usepackage{siunitx}
\usepackage{subfig} 
\usepackage{booktabs}

\usepackage{amsmath,amsfonts}
\usepackage{algorithm}
\usepackage{array}
\usepackage{textcomp}
\usepackage{stfloats}
\usepackage{url}
\usepackage{verbatim}
\usepackage{graphicx}
\usepackage{cite}
\usepackage{booktabs}
\usepackage{threeparttable}  
\usepackage{algpseudocode}

\usepackage{hyperref}
\hypersetup{
    colorlinks=true,
    linkcolor=blue,
    filecolor=magenta,      
    urlcolor=blue,
    citecolor=blue,
}

\usepackage{CJKutf8}

\newcommand{\cov}[1]{\operatorname{cov}\left(#1\right)}

\usepackage{nomencl} 
\usepackage{framed}  
\makenomenclature    

\usepackage{bm} 

\begin{document}


\title{\LARGE \bf RICH-SLAM: Radar SLAM with Incremental and Continuous Hilbert Mapping}


\author{Bingbing Zhang$^{1,2,3}$, Huan Yin$^{3,4}$, Yang Xu$^{3}$, Shuo Liu$^{1,5}$, Shaojie Shen$^{3}$, Fumin Zhang$^{3}$, Wen Xu$^{5,6}$

\thanks{$^{1}$State Key Laboratory of Ocean Sensing, Zhejiang University, China, Email: zhangbb@zju.edu.cn, shuoliu@zju.edu.cn}%
\thanks{$^{2}$Interdisciplinary Student Training Platform for Marine areas, Zhejiang University, Hangzhou, China}%
\thanks{$^{3}$Department of Electronic and Computer Engineering, The Hong Kong University of Science and Technology, Hong Kong, Email: yxuew@connect.ust.hk, eeshaojie@ust.hk, eefumin@ust.hk}
\thanks{$^{4}$School of AI and Robotics, Hunan University, China, Email: huanyin@hnu.edu.cn}
\thanks{$^{5}$Ocean College, Zhejiang University, China, Email: wxu@zju.edu.cn}
\thanks{$^{6}$Institute of Deep-Sea Science and Engineering, Chinese Academy of Sciences, China}%
}

\markboth{Manuscript}%
{Shell \MakeLowercase{\textit{et al.}}: A Sample Article Using IEEEtran.cls for IEEE Journals}


\maketitle

\begin{abstract}
Simultaneous localization and mapping using radar sensors has gained increasing attention due to radar's inherent robustness to adverse weather and lighting conditions. However, radar measurements are characteristically sparse and noisy compared to LiDAR and visual data, posing significant challenges in achieving dense, continuous, and consistent map representations. In this paper, we present RICH-SLAM, a radar SLAM framework designed to address these challenges. Our approach features a Rao-Blackwellized particle filter-based back end that employs particle filtering for pose estimation and Kalman filtering for map updates. We propose an incremental Hilbert-space reduced-rank Gaussian process mapping strategy that enables continuous and uncertainty-aware map representations given sparse radar inputs. We further introduce a posterior-aware particle weighting scheme that leverages the full posterior distribution of map parameters for more robust likelihood evaluation. Experiments on self-collected and public ColoRadar datasets show that RICH-SLAM constructs continuous occupancy maps from sparse radar measurements and supports uncertainty-aware planning for mobile robots.
\end{abstract}


\section*{Supplementary Materials}

\indent A video is submitted as a multimedia attachment. The code and dataset will be released as open source.

\section*{Notation}

Scalars are denoted by italic letters (e.g., $M$, $\ell$), column vectors by lowercase bold letters (e.g., $\bm{p}$, $\bm{x}$), and matrices by uppercase bold letters (e.g., $\bm{H}$, $\bm{K}$). Calligraphic letters (e.g., $\mathcal{GP}$, $\mathcal{T}$, $\mathcal{D}$) are reserved for operators, distributions, or special maps in this study. The bracket $[\cdot]$ indicates a discrete time index, for example, $\bm{x}[i]$ is the vehicle state at step~$i$. Subscripts index spatial samples or basis components (e.g., $\bm{p}_j[i]$, $\phi_j$), while the superscript~$b$ marks body-frame quantities (e.g., $\Delta\bm{p}_j^b$). A hat denotes a posterior estimate (e.g., $\hat{\bm{\theta}}[i]$). Principal symbols are summarized in the adjacent table.

\vspace{4pt}

\textit{Abbreviations}
\begin{IEEEdescription}[\IEEEusemathlabelsep\IEEEsetlabelwidth{Hilbert-GP}]
\item[SLAM]   Simultaneous localization and mapping.
\item[GP]   Gaussian process.
\item[GPOM]   Gaussian process occupancy map.
\item[OGM]    Occupancy grid map.
\item[RBPF]   Rao-Blackwellized particle filter.
\item[SGD]    Stochastic gradient descent.
\item[KF]    Kalman filter.
\item[SoC]    System-on-a-chip (radar).
\item[DoF]    Degrees of freedom.
\item[AUC]    Area under the ROC curve.
\item[RMSE]   Root mean square error.
\item[\textsc{Hilbert-GP}] Hilbert-space reduced-rank Gaussian process.
\item[MAP] Maximum a Posteriori.
\end{IEEEdescription}

\vspace{4pt}

\noindent\textit{Symbols}
\begin{IEEEdescription}[\IEEEusemathlabelsep
    \IEEEsetlabelwidth{$\mathcal{T}(\bm{x},\Delta\bm{p}^b)$}]
\item[$\bm{p}\in\mathbb{R}^2$]
  Spatial query position in world frame.
\item[{$\bm{x}[i]$}]
  Vehicle state $(x,y,\psi)$.
\item[{$\bm{u}[i]$}]
  Odometry (control) input.
\item[{$\bm{v}[i]$}]
  Process noise.
\item[$\mathcal{T}(\bm{x},\Delta\bm{p}^b)$]
  Body-to-world coordinate transform.
\item[$\Delta\bm{p}_j^{b}$]
  $j$-th sampled point in body frame.
\item[$h(\bm{p})$]
  Latent occupancy variable (log-odds) at~$\bm{p}$.
\item[$\mu(\bm{p})$]
  GP mean function (set to zero).
\item[$\kappa(\bm{p},\bm{p}')$]
  GP covariance (kernel) function.
\item[$\sigma_f^2$]
  Signal variance (kernel hyperparameter).
\item[$\sigma_n^2$]
  Measurement noise variance.
\item[$\varsigma(\cdot)$]
  Logistic sigmoid function.
\item[$\ell$]
  Length scale (kernel hyperparameter).
\item[$S(\cdot)$]
  Spectral density of the kernel.
\item[$\phi_j(\bm{p})$]
  $j$-th Laplacian eigenfunction.
\item[$\lambda_j$]
  $j$-th Laplacian eigenvalue.
\item[$\bm{\Lambda}$]
  Spectral density diagonal matrix.
\item[$M$]
  Number of basis functions (features).
\item[$N$]
  Number of sampled points per scan.
\item[$\bm{\Phi}(\bm{p})$]
  Feature vector $[\phi_1,\ldots,\phi_M]^\top\!\in\!\mathbb{R}^{M}$.
\item[{$\bm{H}[i]$}]
  Design matrix $\in\!\mathbb{R}^{M\times N}$.
\item[$\bm{\theta}$]
  Map parameters (feature weights) $\in\!\mathbb{R}^{M}$.
\item[{$\hat{\bm{\theta}}[i]$}]
  Posterior estimate of $\bm{\theta}$.
\item[$\bm{J}$]
  Basis-function index matrix $\in\!\mathbb{N}^{M\times2}$.
\item[$\bm{L}$]
  Half side lengths of the domain.
\item[{$\tilde{\bm{z}}[i]$}]
  Occupancy measurement $\in\!\{-1,+1\}^N$.
\item[{$\bm{z}[i]$}]
  True (latent) log-odds vector $\in\!\mathbb{R}^N$.
\item[{$\hat{\bm{z}}[i]$}]
  Predicted measurement.
\item[$f(\cdot)$]
  State motion model.
\item[{$\mathcal{D}_k[i]$}]
  History and map posterior of particle $k$.
\item[$N_p$]
  Total number of particles.
\item[{$\bm{\Xi}[i]$}]
  Particle set at time step $i$.
\item[{$w_k[i]$}]
  Importance weight of the $k$-th particle.
\item[$k^\star$]
  Index of the largest-weight particle.
\item[$q(\cdot)$]
  Proposal distribution for particle sampling.
\item[$N_{\mathrm{eff}}$]
  Effective sample size.
\item[$T_{\max}$]
  Total number of time steps.
\item[{$\bm{K}[i]$}]
  Kalman gain matrix.
\item[$(\cdot)^\top$]
  Matrix and vector transpose.
\item[$\operatorname{cov}(\cdot)$]
  Covariance matrix.
\item[$\operatorname{var}(\cdot)$]
  Scalar variance.
\item[$\|\cdot\|$]
  Euclidean norm.
\item[$P(\cdot)$]
  Probability density.
\item[$Pr(\cdot)$]
  Probability.
\item[$\mathcal{N}(\bm{\mu},\bm{\Sigma})$]
  Gaussian with mean $\bm{\mu}$, covariance $\bm{\Sigma}$.
\item[$\mathcal{GP}(\mu,\kappa)$]
  GP with mean function $\mu$ and kernel $\kappa$.
\item[$o(\bm{p})$]
  Occupancy probability at query position $\bm{p}$.
\item[$r_e$]
  Endpoint tolerance radius.
\end{IEEEdescription}

\section{INTRODUCTION}

\IEEEPARstart{R}{adar} has emerged as a critical sensor for robotic state estimation due to its robustness in adverse weather conditions~\cite{harlow2024new}. In recent years, radar-based state estimation has gained significant attention; on the other hand, radar sensing is inherently \textit{noisy and sparse} compared to conventionally used LiDAR and visual data. Existing radar-based state estimation methods primarily focus on odometry estimation by integrating radar with other sensing modalities, such as IMUs~\cite{kramer2020radar,huang2024less}, and leveraging radar’s unique physical characteristics, like Doppler velocity measurements~\cite{xu2025incorporating,gentil2025dro}. This combination enhances the robustness and accuracy of odometry, particularly in scenarios where other sensing modalities face limitations.

While radar odometry can provide pose estimation, achieving a fully autonomous system requires mobile robots to have not only accurate state estimation but also \textit{consistent and dense} maps for long-term localization and path planning~\cite{siegwart2011introduction}. This makes SLAM a more suitable choice for radar-based robot navigation compared to odometry-only state estimation. However, building a SLAM system from radar measurements presents significant challenges:
\begin{enumerate}
    \item Robotic navigation systems, particularly those involved in planning, typically require querying the map state at various locations. Radar inherently provides sparse and noisy measurements, causing the map to remain unknown or uncertain across many regions. Consequently, it is highly challenging to construct a map that can be directly employed for robot navigation.
    \item Estimating pose constraints between radar measurements is more challenging compared to LiDAR. Traditional point registration~\cite{pomerleau2015review} or feature association~\cite{qin2018vins} are {less effective} for radar-based state estimation~\cite{kubelka2024we}. Sparse radar beams provide far fewer constraints for pose estimation, leading to multi-modal posterior position estimates, which hinders the use of mainstream pose-graph optimization for dense mapping~\cite{grisetti2011tutorial}.
\end{enumerate}

Early range sensing sensors, such as 2D laser scanners, faced similar challenges in providing sparse and noisy measurements. To address these issues, researchers proposed using particle filter frameworks for pose estimation, as an alternative to parametric-based filter and optimization methods. For example, FastSLAM~\cite{montemerlo2003fastslam} employed a particle filter in the back end to estimate robot poses while simultaneously building a map. A classical example of FastSLAM is GMapping~\cite{grisetti2005improving,grisetti2007improved}, which used occupancy grid maps (OGM) as the basis for grid-based mapping. However, the sparse nature of radar data introduces challenges in dense mapping, particularly as it results in many grid cells being unable to be updated on resource-constrained platforms. Additional limitations of OGM are detailed in Section~\ref{sec:related}.

In this study, we propose to leverage \textit{Gaussian Process}-based representations~\cite{o2009contextual}, which model the correlations between grid cells, rather than relying on conventional fixed-size grid maps. On the other hand, Gaussian Process occupancy maps (GPOM) are computationally expensive for both update and prediction operations. To address this issue, inspired by the work proposed by Ramos and Ott~\cite{ramos2016hilbert}, we employ Hilbert space reduced rank Gaussian process (\textsc{Hilbert-GP})~\cite{solin2020hilbert}, which enhances Gaussian process efficiency while maintaining dense map representations. The \textsc{Hilbert-GP} map also provides uncertainty prediction, which distinguishes it from the original method introduced in ~\cite{ramos2016hilbert}. At the back end, we employ a particle filter framework for state estimation, which captures the multi-modal pose posteriors induced by sparse radar returns.

In contrast to particle weighting schemes~\cite{hata2016particle,vallicrosa2018h} that ignore map uncertainty, our method leverages the full posterior distribution of map parameters to calculate particle likelihoods, resulting in a posterior-aware weighting mechanism for sparse radar observations. Since even a small rotation misaligns all sampled radar points with the continuous occupancy field simultaneously, this map-based weighting could be informative for constraining heading. In addition, sparse radar endpoints are corrupted by range-angle noise and specular reflection and thus should not be treated as exact geometric hits, we further make the likelihood endpoint-tolerant over a small neighborhood around each endpoint. Together, these components enable a \textit{radar SLAM system that uses sparse radar signals to build dense and continuous maps while reducing the localization error}.

In summary, we present RICH-SLAM, a radar SLAM system  built on incremental and continuous Hilbert mapping. Our contributions are threefold:
\begin{itemize}
    \item We adapt the Hilbert-space Gaussian process to sparse radar beams, yielding an incremental mapping module that produces continuous occupancy maps. The produced map has posterior uncertainty and supports resolution-independent querying at arbitrary locations.

    \item Within a Rao-Blackwellized particle filter, we couple the continuous Hilbert map with a posterior- and endpoint-aware particle weighting scheme.
    
    \item We demonstrate that the learned continuous occupancy and uncertainty fields can be used as traversability costs for mobile robot planning.
\end{itemize}
\par Each contribution above is validated by a set of experiments. We use a fixed basis size, length scale, radar range, and particle count across all sequences, while only changing the signal and measurement variances and the Hilbert domain due to the environment and sensor differences.



\section{Related Work}
\label{sec:related}

We first introduce the radar-based SLAM in Section~\ref{sec:radar}, and then discuss the occupancy mapping using range sensors in Section~\ref{sec:occupancy}. 

\begin{table*}[t]
\caption{Comparison of Different Map Representations}
\centering
\renewcommand{\arraystretch}{1.2} 
\resizebox{\textwidth}{!}{ 
\begin{threeparttable}
\begin{tabular}{ccccc}
 \hline
 \hline
 \textbf{ } & \textbf{OGM} \cite{30720,thrun2003learning} &\textbf{GPOM} \cite{o2012gaussian}& \textbf{Hilbert map} \cite{ramos2016hilbert} & \textbf{Hilbert-GP map} (This Work)\\ 
 \hline
 Representation&Parametric&Non-parametric&Parametric&Parametric\\
 \hline 
\multirow{2}*{Assumption}&Bernoulli &Latent state  & Latent state follows&Latent state follows\\
&{distribution}&follows Gaussian process&  Gaussian process&Gaussian process\\
  \hline
Continuous?&No&Yes&Yes&Yes\\
\hline
\multirow{3}*{Hyper parameters} &Log-odds     &Length scale& Regularization strength&Length scale \\
 &Cell resolution&  Signal variance& Kernel width&Signal variance\\
& &Measurement noise variance &$L_1$ ratio &Measurement noise variance\\
\hline
Uncertainty derivable?&No&Yes&No&Yes\\
\hline
Inference cost (per point)&$\mathcal{O}(1)$&$\mathcal{O}(N_{\mathrm{tr}}^3)$&$\mathcal{O}(M)$&$\mathcal{O}(M)$ for mean $\mathcal{O}(M^2)$ for covariance\\
\hline
Learning cost (per iteration)&$\mathcal{O}(1)$&$\mathcal{O}(N_{\mathrm{tr}}^3)$&$\mathcal{O}(M)$&$\mathcal{O}(M^2)$\\
 \hline
 \hline
\end{tabular}
\begin{tablenotes}
\item {\footnotesize \dag~ $N_{\mathrm{tr}}$ is the total number of points, $M$ is the number of features.}
\end{tablenotes}
\end{threeparttable}
} 
\label{tabCompare}
\end{table*}

\subsection{Radar-based SLAM}
\label{sec:radar}

Radar sensors in robotics can be categorized into two types~\cite{carlone2025slam}: spinning radar and system-on-a-chip (SoC) radar. Spinning radars, such as the Navtech, rotate mechanically to provide dense 360-degree range-azimuth images with high angular resolution. These sensors have been employed in large-scale state estimation systems~\cite{yin2021rall, hong2022radarslam}, which demonstrated robust localization and mapping in all weather conditions. However, spinning radars are expensive and typically do not provide Doppler velocity information~\cite{lisus2024doppler}. In this study, RICH-SLAM is developed using SoC radar, which features a lightweight design and reduced power requirements compared to spinning radar. However, SoC radar presents challenges for state estimation due to its sparse and noisy measurements.

To enhance robustness and accuracy, recent approaches have integrated inertial data for radar-based state estimation. Early efforts primarily adopted filtering frameworks: Doer and Trommer~\cite{doer2020ekf} implemented an extended Kalman filter fusing Doppler velocity with IMU propagation, Kramer \textit{et al.}~\cite{kramer2020radar} proposed sliding-window optimization leveraging Doppler velocity, and Michalczyk \textit{et al.}~\cite{michalczyk2023multi} developed a tightly-coupled multi-state EKF incorporating radar landmarks, where persistent landmarks with high radar cross-section were tracked across frames to provide constraints.

While filtering-based methods offer computational efficiency, scan registration techniques can exploit the full geometric structure of radar scans. The 4D iRIOM~\cite{zhuang20234d} employed Doppler-based noise filtering and distribution-to-multi-distribution matching, closely aligning with LiDAR-based state estimation techniques. Zhang \textit{et al.}~\cite{4DRadarSLAM} proposed 4DRadarSLAM, which performs scan-to-scan matching based on a probabilistic GICP and employs intensity scan context for loop closure within a pose-graph optimization framework. Wang \textit{et al.}~\cite{wang2024riv} presented RIV-SLAM, which jointly optimizes the pose and ego velocity by tightly coupling radar data with IMU pre-integration, incorporating ground extraction and a registration approach tailored for anisotropic radar measurements. Casado-Herraez \textit{et al.}~\cite{casado2025rai} proposed RaI-SLAM, a modular radar-inertial approach that exploits velocity and radar cross-section information and corrects accumulated drift through a global pose graph with loop closures.

Beyond model-based estimation, Lu \textit{et al.}~\cite{lu2020milliego} introduced a data-driven approach, using convolutional neural networks to encode radar data and recurrent neural networks to process IMU measurements for trajectory estimation. Despite the diversity of these methods, most of them regard radar data as mere 3D points, often lacking analysis of measurement uncertainty. To address this gap, Kellner \textit{et al.}~\cite{kellner2014instantaneous} considered angular uncertainty in Doppler residuals, while Xu \textit{et al.}~\cite{xu2025incorporating} proposed a comprehensive sensor fusion framework that incorporates full point uncertainty estimation, modeling the noise characteristics inherent in radar measurements.

Regardless of the odometry strategy employed, pose estimates inevitably drift over time, making loop closure critical for constructing globally consistent maps~\cite{yin2024survey}. Early radar loop closure methods relied on handcrafted descriptors: Holder \textit{et al.}~\cite{holder2019real} proposed detecting loops by applying landmark relations with iterative closest point (ICP) for relative pose estimation, existing studies~\cite{hong2020radarslam,hong2022radarslam} adapted the LiDAR-oriented descriptor M2DP~\cite{he2016m2dp} to the 2D radar point cloud, and 4D iRIOM~\cite{zhuang20234d} integrated Scan Context~\cite{kim2018scan} with generalized ICP~\cite{koide2021voxelized} for loop closure. Given the sparsity of radar scans, several works~\cite{li20234d,4DRadarSLAM,wang2024riv} further adapted intensity scan context~\cite{wang2020intensity} to encode radar point clouds more effectively. However, handcrafted descriptors may struggle to capture complex scene geometry, motivating learning-based alternatives. Meiresone \textit{et al.}~\cite{meiresone2024loop} proposed an autoencoder-based method for place recognition with low-resolution single-chip mmWave radar, demonstrating feasibility in indoor environments without relying on ground truth poses. Usuelli \textit{et al.}~\cite{usuelli2023radarlcd} introduced RadarLCD, a supervised deep learning pipeline leveraging pre-trained radar odometry features for key point selection. More recently, Peng \textit{et al.}~\cite{peng2024transloc4d} proposed TransLoc4D, a transformer-based approach for radar place recognition, while Cheng \textit{et al.}~\cite{cheng2025radarpr} introduced a context-aware method that captures multi-scale local features for robust place recognition in harsh scenarios. Note that all the methods above do not focus on the dense and continuous representations for radar mapping.

A comprehensive review of radar-based state estimation is beyond the scope of this paper. We recommend readers refer to the recent survey by Harlow \textit{et al.}~\cite{harlow2024new} for an in-depth overview of radar in robotics.

\subsection{Occupancy Mapping}
\label{sec:occupancy}

The radar-based SLAM methods above primarily address the localization component. We now turn to occupancy mapping and the complementary challenge of environment representation that RICH-SLAM jointly addresses.

OGM~\cite{30720,thrun2003learning} represent one of the most foundational and widely adopted environmental representations for mobile robots. In OGM, the environment is discretized into a grid of cells, with each cell modeling the occupancy state as an independent binary random variable. This discrete, probabilistic framework has been extensively applied across a wide range of robotic tasks, including localization and mapping~\cite{hess2016real}, as well as full navigation systems~\cite{sodhi2019online,ren2024rog,banfi2022worth}, establishing itself as a classic and reliable map.

On the other hand, despite their widespread adoption in the mobile robotics community, OGM have notable limitations. First, they rely on the assumption of independence between cells, neglecting the inherent spatial dependencies of real-world environments. This results in high uncertainty in occluded regions and sensor beam gaps. Second, OGM are restricted to a single fixed resolution, making robots suffer from discretization errors and require significant memory for detailed environmental mapping. Third, conventional occupancy mapping techniques often struggle to handle sparse, noisy, multi-source sensor data. These limitations have motivated the development of continuous representations for OGM.

One of the most notable advancements in this domain is GPOM~\cite{o2012gaussian}, which models occupancy probability as a continuous function over space using Gaussian processes as a non-parametric Bayesian learning technique. This continuous formulation eliminates discretization artifacts, ensures smoothness, and enables representation of the environment at arbitrary resolutions. Subsequent works extended GPOM in several directions: contextual GPOM~\cite{ocallaghan2010contextual} incorporated sensor and location uncertainty into the GP framework, while Kim and Kim~\cite{kim2022mixture} proposed a mixture of Gaussian processes approach that clusters observations into manageable subsets, dramatically reducing the computational complexity while preserving local map structures. However, the computational burden associated with standard GPOM remains significant, as the cubic complexity in the number of data points makes it impractical for real-time, large-scale applications. To provide a clearer understanding, Table~\ref{tabCompare} summarizes the principal characteristics that distinguish OGM, GPOM, and Hilbert Maps. The \textsc{Hilbert-GP}-based map in the proposed RICH-SLAM system serves as an
alternative to GPOM; however, unlike Hilbert Maps, it is designed to predict the mapping uncertainty. This feature could benefit applications such as improving state estimation and deploying path finding methods.

The work most closely related to ours is H-SLAM~\cite{vallicrosa2018h},  which also embeds Hilbert maps within a Rao-Blackwellized particle filter. RICH-SLAM differs in three key aspects. First, H-SLAM adopts the original Hilbert map~\cite{ramos2016hilbert} with deterministic weights optimized by SGD, providing only point estimates of occupancy; RICH-SLAM instead builds on the Hilbert-space Gaussian process formulation~\cite{solin2020hilbert}, which maintains a full posterior over map parameters and yields principled uncertainty. Second, RICH-SLAM replaces SGD updates with a closed-form Kalman update, eliminating learning-rate tuning and improving robustness under noisy measurements. Third, particle weights in RICH-SLAM are computed from the full map posterior rather than its mean alone, which is critical for the sparse and noisy returns of millimeter-wave radar.

More recently, neural implicit representations have emerged as a powerful paradigm for continuous map construction in SLAM. Methods such as iMAP~\cite{sucar2021imap}, NICE-SLAM~\cite{zhu2022nice}, and PIN-SLAM~\cite{pan2024pin} represent the scene as continuous neural fields, enabling high-fidelity surface reconstruction and compact map storage. PIN-SLAM~\cite{pan2024pin}, for instance, uses sparse optimizable neural points to build globally consistent maps that are inherently elastic and deformable during loop closure. While these neural implicit approaches have shown remarkable results for visual and LiDAR SLAM, they typically require dense observations for effective training, and their applicability to extremely sparse radar returns remains largely unexplored. In contrast, the \textsc{Hilbert-GP} representation adopted in RICH-SLAM is well suited to the sparse measurement input, offers uncertainty estimates, and requires no GPU-based training, making it practical for resource-constrained robotic platforms.

\section{Preliminaries}
\label{sec:preliminaries}

This section introduces the foundational concepts in RICH-SLAM. Section~\ref{sec:problem} states the 2D mapping assumption adopted in this work. Section~\ref{sec:gpom} presents Gaussian process occupancy maps, and Section~\ref{sec:hilbert} describes the Hilbert space approximation employed for computational efficiency.

\begin{figure*}[t]
    \centering
    \includegraphics[width=0.98\linewidth]{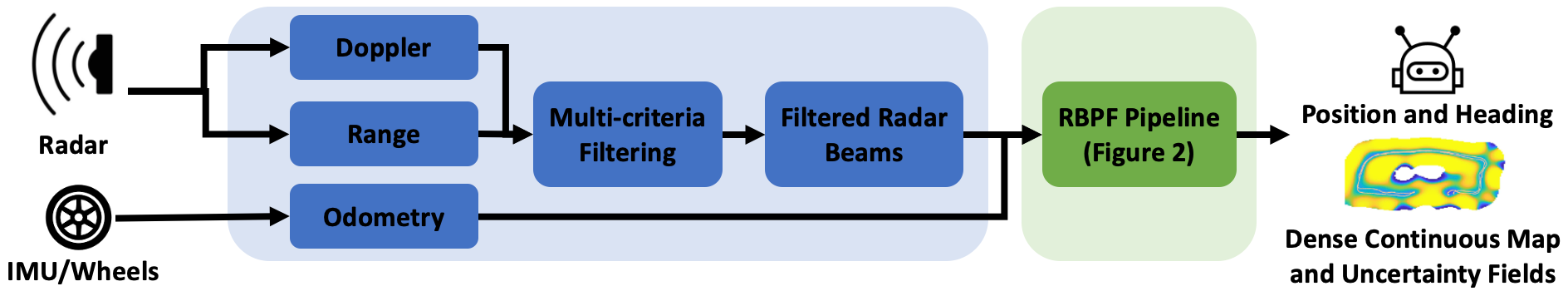}
    \caption{Overview of the RICH-SLAM framework. Radar Doppler and range measurements pass through a multi-criteria front-end filter  that rejects outliers and outputs a set of filtered radar beams. The filtered beams and the odometry are fed into the Rao-Blackwellized particle filter back end, which jointly estimates the robot pose and a per-particle continuous occupancy map. The framework outputs the position and heading together with a dense continuous occupancy map and its predictive uncertainty field.}
    \label{fig:Rbpf}
\end{figure*}

\subsection{Problem Statement}
\label{sec:problem}

The goal of RICH-SLAM is to jointly estimate the robot trajectory and a continuous occupancy map from sequential radar measurements and odometry inputs. Denoting the robot pose at time step $i$ as $\bm{x}[i]$, the map parameters as $\bm{\theta}$, the radar observations as $\tilde{\bm{z}}[1:i]$, and the odometry inputs as $\bm{u}[1:i]$, the problem is formulated as estimating the posterior distribution:
\begin{equation}
    P\bigl(\bm{x}[1:i],\,\bm{\theta}\;\big|\;\tilde{\bm{z}}[1:i],\,\bm{u}[1:i]\bigr)
\end{equation}
where the map is represented as a continuous function parameterized by $\bm{\theta}$ through the Hilbert space approximation introduced in Section~\ref{sec:hilbert}. The specific factorization and inference strategy for this posterior are detailed in Section~\ref{sec:rich}.

In this study, we assume that radar SLAM is performed in a 2D world with a 3-DoF pose. Height estimation is excluded for the following reasons:
\begin{itemize}
    \item Radar-based state estimation is widely used for land vehicles on planar surfaces~\cite{huang2024less,zhuang20234d}, where height estimation is typically unnecessary, as horizontal distance and velocity are prioritized over vertical dimensions.
    \item The lower angular accuracy of radar compared to lidar renders height estimation unreliable, and focusing on 2D mapping simplifies the problem without compromising accuracy.
    \item Height variations can be addressed through global optimization techniques, such as barometric sensors for aerial vehicles or pressure sensors for underwater vehicles~\cite{kohlbrecher2011flexible}.
\end{itemize}
\par Therefore, focusing on a 2D map is both practical and effective in the context of land vehicle radar SLAM. Reducing the state dimension in SLAM systems is a common approach in various works, such as~\cite{qin2018vins,kohlbrecher2011flexible}. This reduction simplifies the problem while retaining sufficient accuracy for practical applications. Extending to 3D (6-DoF) radar SLAM is discussed in Section~\ref{sec:conclusion}.

\subsection{Gaussian process occupancy maps}
\label{sec:gpom}

To model the continuous occupancy map of this study, we place a GP prior~\cite{o2012gaussian} over the log-odds occupancy function $h\colon \mathbb{R}^{2}\!\to\!\mathbb{R}$:
\begin{equation}
    \label{eq:GPOM}
    h(\bm{p}) \sim \mathcal{GP}\bigl(\mu(\bm{p}),\,
    \kappa(\bm{p}, \bm{p}')\bigr)
\end{equation}
where $h(\bm{p})$ is a latent variable encoding the occupancy probability at position $\bm{p}$ in a log-odds representation; $\mu(\cdot)$ is the mean function, initialized to zero to reflect an uninformative prior; and $\kappa(\cdot,\cdot)$ is a covariance kernel encoding spatial correlations. We adopt the squared exponential (SE) kernel:
\begin{equation}
    \label{eqSEKernel}
    \kappa_\mathrm{SE}(\bm{p}, \bm{p}') = \sigma_f^2
    \exp\!\left(-\frac{\|\bm{p} - \bm{p}'\|^2}
    {2\,\ell^2}\right)
\end{equation}
where $\sigma_f^2 > 0$ (signal variance) governs the covariance magnitude, $\ell > 0$ (length scale) controls the smoothness, and $\|\cdot\|$ denotes the Euclidean norm.

Given a set of occupancy labels $\tilde{z}(\bm{p}) \in \{-1, +1\}$, where $-1$ and $+1$ indicate free and occupied respectively (detailed in Figure~\ref{fig:BeamSampling}), the observation likelihood is modeled as a Bernoulli distribution~\cite{bishop2006pattern} through the sigmoid function $\varsigma(\cdot)$:
\begin{equation}
    \label{eq:Prob}
    Pr(\tilde{z}=1\,|\,\bm{p}) = \varsigma\bigl(h(\bm{p})\bigr)
\end{equation}

The GP formulation above provides a principled probabilistic framework for continuous occupancy mapping. However, exact GP inference incurs $\mathcal{O}(n^{3})$ computational cost in the number of observations~\cite{rasmussen2006gaussian}. In the following section, we introduce a reduced-rank approximation that parameterizes the map by a finite-dimensional weight vector $\bm{\theta}\in\mathbb{R}^{M}$.



\subsection{Hilbert Space Approximation}
\label{sec:hilbert}

Given the computational complexity of standard GPs, we employ \textsc{Hilbert-GP}~\cite{solin2020hilbert} to approximate the covariance function through an eigenfunction expansion of the Laplace operator on a bounded domain:
\begin{equation}
\label{eqDecomposition}
\kappa(\bm{p},\bm{p}') \approx \sum_{j=1}^{M} S\!\left(\sqrt{\lambda_j}\right)\phi_j(\bm{p})\,\phi_j(\bm{p}')
= \bm{\Phi}(\bm{p})^{\top}\bm{\Lambda}\,\bm{\Phi}(\bm{p}')
\end{equation}
where $\phi_j(\bm{p})$ are orthonormal eigenfunctions with corresponding eigenvalues $\lambda_j$, which serve as the \emph{basis functions} (also called features) in our parametric approximation, and $S(\cdot)$ denotes the kernel's spectral density.

To obtain closed-form eigenfunctions, we define a compact rectangular domain $\Omega = [-\bm{L}_1, \bm{L}_1] \times [-\bm{L}_2, \bm{L}_2]$ centered at the origin, where $\bm{L} = [\bm{L}_1, \bm{L}_2]^\top$ collects the half-side-lengths. On $\Omega$, the eigenfunctions and eigenvalues are given by
\begin{equation}
\begin{aligned}
\phi_j(\bm{p}) & =\prod_{d=1}^2 \frac{1}{\sqrt{\bm{L}_d}} \sin \left(\frac{\pi \bm{J}_{j, d}\left(p_d+\bm{L}_d\right)}{2 \bm{L}_d}\right) \\
\lambda_j & =\sum_{d=1}^2\left(\frac{\pi \bm{J}_{j, d}}{2 \bm{L}_d}\right)^2
\end{aligned}
\label{eqHilbertGPBasis}
\end{equation}
where $\bm{J}_{j,d}$ denotes the $(j,d)$-th entry of the index matrix $\bm{J}$. The index matrix $\bm{J}$ enumerates all two-dimensional frequency pairs constructed from the integers $1, 2, \ldots, \sqrt{M}$, with $M$ constrained to be a perfect square:
\begin{equation}
\bm{J}^\top = \begin{bmatrix} 1 & 1 & \cdots & 1 & 2 & \cdots & \sqrt{M} \\ 1 & 2 & \cdots & \sqrt{M} & 1 & \cdots & \sqrt{M} \end{bmatrix}
\end{equation}
\par For the squared exponential kernel in~\eqref{eqSEKernel}, the spectral density evaluates to
\begin{equation}
S(\sqrt{\lambda_j})=\sigma_f^2(2 \pi \ell^2) \exp \left(-\frac{\lambda_j \ell^2}{2}\right)
\end{equation}
where $\sigma_f^2$ and $\ell$ are defined in Equation~\eqref{eqSEKernel}.

Then, substituting Equation~\eqref{eqDecomposition} into the GP model can yield a finite-dimensional parametric form for the latent occupancy field:
\begin{equation}
    \label{eq:Latent}
    h(\bm{p}) \approx \bm{\Phi}\left(\bm{p}\right)^\top \bm{\theta}
\end{equation}
where $\bm{\Phi}\left(\bm{p}\right)=\left[\phi_1\left(\bm{p}\right) \; \phi_2\left(\bm{p}\right) \cdots \phi_M\left(\bm{p}\right)\right]^\top$ maps the position $\bm{p}$ to a high-dimensional feature vector, and $\bm{\theta} \in \mathbb{R}^M$ is the weight vector to be estimated. Under the GP prior, $\bm{\theta}$ has zero mean and diagonal covariance:
\begin{equation}
\mathrm{cov}\!\left(\hat{\bm{\theta}}[0]\right)
= \mathrm{diag}\!\left(S(\sqrt{\lambda_1}),\ldots,S(\sqrt{\lambda_M})\right)
= \bm{\Lambda}
\label{eq:cov}
\end{equation}

Equation~\eqref{eq:Latent} decomposes the map into a fixed set of basis functions $\bm{\Phi}(\bm{p})$ and a weight vector $\bm{\theta}$ that encodes spatial occupancy information. In the context of SLAM, $\bm{\theta}$ is the central quantity to be estimated: as the robot traverses the environment and collects radar measurements, the posterior distribution of $\bm{\theta}$ is updated incrementally. The linear relationship between $h(\bm{p})$ and $\bm{\theta}$ enables closed-form Bayesian updates, as detailed in Section~\ref{sec:update}, while the prior covariance in Equation~\eqref{eq:cov} captures the initial map uncertainty before any measurement is incorporated.

Concretely, learning the map reduces to maintaining a single Gaussian belief over the weight vector $\bm{\theta}$. The basis $\bm{\Phi}(\bm{p})$ is fixed once the kernel and domain are chosen and is never updated; all map information is carried by the posterior over $\bm{\theta}$. Its mean $\hat{\bm{\theta}}$ defines the current occupancy estimate through $h(\bm{p})=\bm{\Phi}(\bm{p})^{\top}\hat{\bm{\theta}}$, whereas its covariance $\mathrm{cov}(\hat{\bm{\theta}})$ defines the map uncertainty at any query point through $\bm{\Phi}(\bm{p})^{\top}\mathrm{cov}(\hat{\bm{\theta}})\,\bm{\Phi}(\bm{p})$. As each radar scan arrives, this closed-form update moves $\hat{\bm{\theta}}$ to better explain the new occupancy samples and shrinks $\mathrm{cov}(\hat{\bm{\theta}})$ in the observed region. Equivalently, this Gaussian belief over $\bm{\theta}$ is the weight-space form of the GP posterior over the occupancy field.

Beyond enabling this recursive estimation, the parameterization offers two practical properties. First, since the map is a continuous function of $\bm{p}$, occupancy can be queried at arbitrary locations and resolutions after learning, without retraining or changing the stored representation. Second, the Gaussian posterior over $\bm{\theta}$ yields a closed-form predictive uncertainty at every query point. Both properties are demonstrated in the experimental sections.


\section{RICH-SLAM}
\label{sec:rich}

\begin{figure*}[t]
    \centering
    \includegraphics[width=0.98\linewidth]{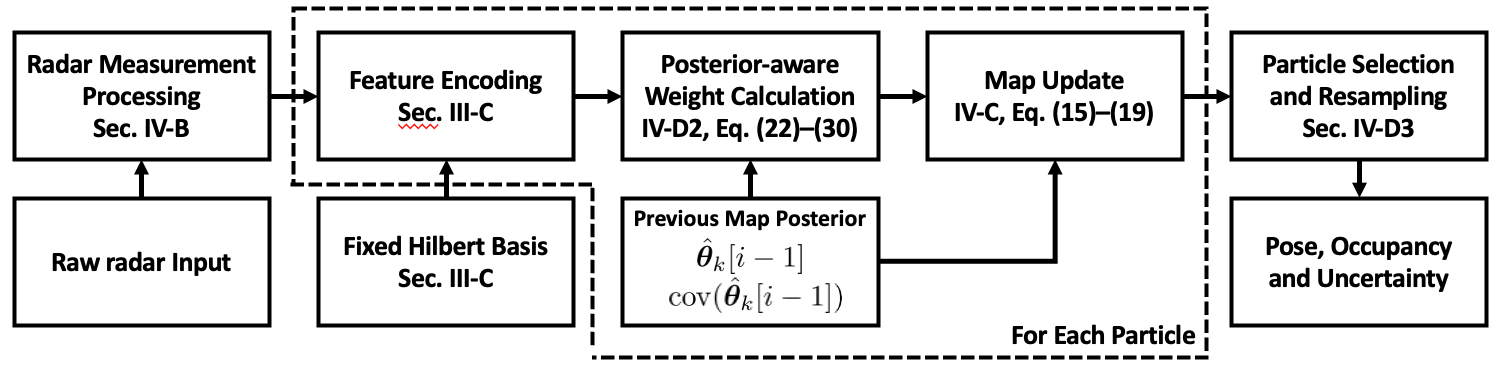}
    \caption{Radar computation flow within one RBPF iteration at time step $i$. The fixed Hilbert basis is shared across all particles, whereas each particle maintains its own map posterior (dashed box), updated by the Kalman filter and reused as the prior at step $i{+}1$. The odometry-driven particle propagation is omitted for clarity.}
    \label{fig:flow}
\end{figure*}

This section presents the proposed RICH-SLAM framework in detail. Figure~\ref{fig:Rbpf} illustrates the overall system architecture, which consists of three main components: a multi-criteria front-end filter for radar beam processing, a Rao-Blackwellized particle filter for pose estimation, and a Kalman filter-based map update module. Figure~\ref{fig:flow} summarizes the computation flow within one RBPF iteration.

Our presentation follows the RBPF factorization stated in Section~\ref{sec:estimation}. After motivating the state estimation design (Section~\ref{sec:estimation}) and specifying the shared radar measurement model (Section~\ref{sec:Processing}), we develop the two factors in turn: the conditional map posterior $P(\bm{\theta}\mid\bm{x},\tilde{\bm{z}})$ via a closed-form Kalman update (Section~\ref{sec:update}), and the pose posterior $P(\bm{x}\mid\tilde{\bm{z}},\bm{u})$ via a particle filter (Section~\ref{sec:pf}). Section~\ref{sec:pipeline} then assembles both factors into the complete RICH-SLAM loop.

\subsection{State Estimation Design}
\label{sec:estimation}

Recall from Section~\ref{sec:problem} that the central inference task is to compute the posterior $P\bigl(\bm{x}[1:i],\,\bm{\theta} \mid \tilde{\bm{z}}[1:i],\,\bm{u}[1:i]\bigr)$. Our design choice is driven by two practical challenges of radar-based SLAM.

The first challenge arises from \emph{pose estimation}. The sparsity of radar returns frequently causes perceptual aliasing, where measurements from geometrically distinct locations appear similar. This gives rise to multimodal posterior distributions over the robot pose. Kalman filters, which generally assume a unimodal Gaussian, cannot capture such multi-modality. This motivates the use of a \emph{particle filter} in this study, a non-parametric approach that represents the pose posterior with a weighted set of hypotheses, each assigned an importance weight reflecting its likelihood given the observations (detailed in Section~\ref{sec:pf}).

The second challenge concerns \emph{map estimation}. Ideally, the particle filter would sample the joint pose and map state; however, because the map parameter $\bm{\theta}\!\in\!\mathbb{R}^{M}$ is high-dimensional (typically $M>100$), the number of particles needed to cover the joint space grows exponentially with the state dimension~\cite{durrant2006simultaneous}, making direct sampling intractable. To address this, we adopt an RBPF~\cite{montemerlo2003fastslam}, which factorizes the posterior as
\begin{equation}\label{eq:rbpf}
\begin{aligned}
  &P\bigl(\bm{x}[1\!:\!i],\,\bm{\theta}
         \mid \tilde{\bm{z}}[1\!:\!i],\,\bm{u}[1\!:\!i]\bigr) \\
  &\quad=
  \underbrace{P\bigl(\bm{\theta}
         \mid \bm{x}[1\!:\!i],\,
              \tilde{\bm{z}}[1\!:\!i]\bigr)}_{\text{map update}}
  \;\;
  \underbrace{P\bigl(\bm{x}[1\!:\!i]
         \mid \tilde{\bm{z}}[1\!:\!i],\,
              \bm{u}[1\!:\!i]\bigr)}_{\text{pose estimation}}
\end{aligned}
\end{equation}
so that particles are maintained only for the low-dimensional pose trajectory, while the map is estimated \emph{conditionally} on each particle's trajectory. Crucially, because the latent map model is linear in~$\bm{\theta}$ (see Equation~\eqref{eq:Latent}), the conditional map posterior admits a closed-form Gaussian representation and can be updated via a Kalman filter (detailed in Section~\ref{sec:update}). Compared with the sampling-based map inference in~\cite{wigren2019parameter}, this closed-form solution reduces computational cost and provides analytical uncertainty estimates for the map parameters.

These two factors structure the remainder of this section, each resolving one of the challenges above. The first factor, the conditional map posterior, is developed in Section~\ref{sec:update}, where $\bm{\theta}$ is handled by a closed-form Kalman update rather than by sampling. The second factor, the pose posterior, is developed in Section~\ref{sec:pf}, where the multimodality of the pose is handled by the particle filter. Both factors rely on the shared radar measurement model introduced as follows.

\begin{figure}
    \centering
    \includegraphics[width=0.99\linewidth]{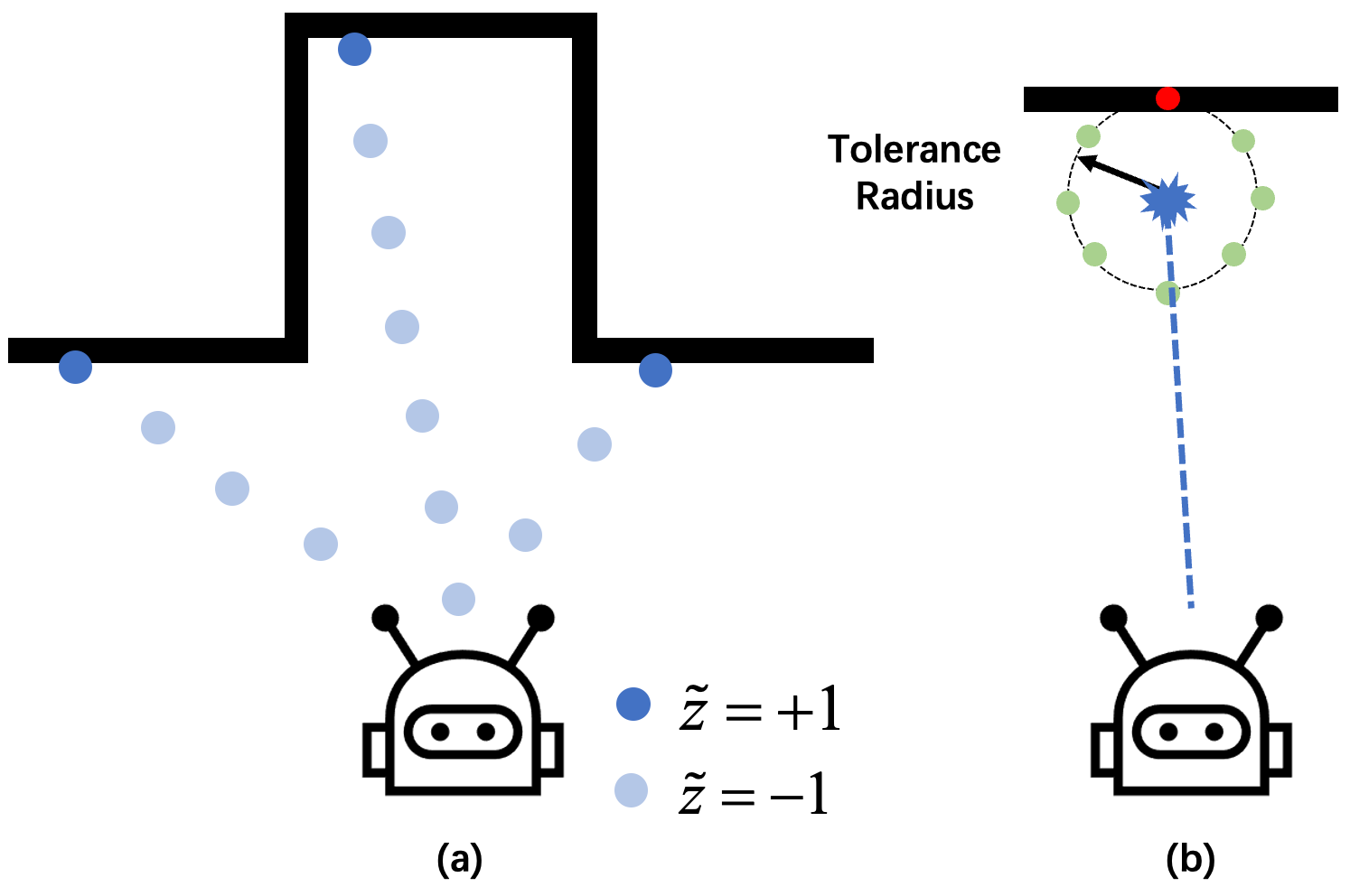}
    \caption{(a)~Sampling of the occupancy field along filtered radar beams: free-space samples are labeled $\tilde z=-1$ (light blue) and beam endpoints $\tilde z=+1$ (dark blue), interpreted as measured log-odds consistent with Equation~\eqref{eq:GPOM}. (b)~Endpoint tolerance (Section~\ref{sec:uncertainty}): instead of querying only the reported endpoint (blue star), the occupancy field is evaluated at neighboring points (green) within a tolerance radius $r_e$, and the maximum occupancy (red) among them (Equation~\eqref{eq:endpoint_tolerance}) is taken as the endpoint likelihood.}
    \label{fig:BeamSampling}
\end{figure}

\subsection{Radar Measurement Processing}
\label{sec:Processing}

Building on the framework above, we now describe the concrete system components. The inputs consist of two parts: a vehicular odometry-based motion model and a radar-based measurement model. For the odometry, we omit explicit dependence on the dynamic model or specific sensor measurements for notational simplicity and generality; potential sources include IMU/wheel measurements~\cite{niu2021wheel}, radar inertial odometry~\cite{huang2024less,doer2020ekf,xu2025incorporating} and others. The remainder of this subsection focuses on the radar measurement model.

Radar measurements inherently contain far more outliers than LiDAR or camera data, which can significantly degrade state estimation performance. These outliers arise primarily from noise and multi-path effects. Valid beams, however, can typically be matched across consecutive frames~\cite{huang2024less}, a property that motivates a multi-frame tracking strategy for outlier rejection.

Specifically, following the filtering criteria in~\cite{xu2025incorporating}, we retain only beam points that can be successfully tracked over multiple consecutive frames. A beam point is deemed reliably tracked if it passes filters on radar point velocity, intensity, and modeled uncertainty. Only repeatable points observed across multiple frames are accepted as reliable measurements and forwarded to the state estimation module.

Given the filtered beams, we construct a measurement model that samples multiple points along each radar beam, as illustrated in Figure~\ref{fig:BeamSampling}. This extends the beam model in~\cite{xu2025incorporating}, where only a single point per beam was used. The measurement values, either $-1$ or $+1$, are interpreted as the ``measured'' log odds of the occupancy probability, ensuring consistency with Equation~\eqref{eq:GPOM}.

Given a sampled point $\bm{p}_n^w[i]$ in the world frame~$w$, the per-point likelihood follows from Equation~\eqref{eq:Prob}:
\begin{equation}
\small
  \label{eq:Bernoulli}
  Pr\!\left((\tilde{\bm{z}}[i])_n
            \mid (\bm{z}[i])_n\right)
  =
  \begin{cases}
    \varsigma\!\bigl((\bm{z}[i])_n\bigr),
      & \text{if } (\tilde{\bm{z}}[i])_n = 1 \\[4pt]
    1 - \varsigma\!\bigl((\bm{z}[i])_n\bigr),
      & \text{if } (\tilde{\bm{z}}[i])_n = -1
  \end{cases}
\end{equation}
where $\tilde{\bm{z}}[i]$ is the length-$N$ measurement vector at time step~$i$, and $\bm{z}[i]$ is the corresponding true latent state. The latent state can be approximated by evaluating the map function at each sampled point:
\begin{equation}
  \label{eq:predictedMeasurement}
  \bm{z}[i]
  \approx \bigl[h(\bm{p}_n^w[i])\bigr]_{n=1}^{N}
  =
  \begin{bmatrix}
    h(\bm{p})\big|_{\bm{p}
      =\mathcal{T}(\bm{x}[i],\,\Delta\bm{p}_1^b)} \\[2pt]
    h(\bm{p})\big|_{\bm{p}
      =\mathcal{T}(\bm{x}[i],\,\Delta\bm{p}_2^b)} \\
    \vdots \\
    h(\bm{p})\big|_{\bm{p}
      =\mathcal{T}(\bm{x}[i],\,\Delta\bm{p}_N^b)}
  \end{bmatrix}
\end{equation}
where $\Delta\bm{p}_n^b$ is the relative position of the $n$-th sampled point in the body frame, and $\mathcal{T}(\bm{x}[i],\Delta\bm{p}_n^b)$ transforms it into the world frame using the current pose $\bm{x}[i]$. Each element on the right-hand side is computed via Equation~\eqref{eq:Latent}. For notational brevity, we omit the superscript $w$ in the remainder of this paper; all sampled positions $\bm{p}_n[i]$ (and the particle-indexed $\bm{p}_{k,n}[i]$) are expressed in the world frame.

Assuming conditional independence across the $N$ sampled points, the joint measurement likelihood becomes:
\begin{equation}
  \label{eq:beams}
  Pr\!\bigl(\tilde{\bm{z}}[i]\mid\bm{z}[i]\bigr)
  = \prod_{n=1}^{N}
    Pr\!\left((\tilde{\bm{z}}[i])_n
              \mid (\bm{z}[i])_n\right)
\end{equation}
\par Equations~\eqref{eq:Bernoulli} and~\eqref{eq:beams} define the likelihood under the assumption that the true latent state is known. In practice, however, the latent state is uncertain. In Section~\ref{sec:uncertainty}, we extend this formulation by marginalizing over the posterior distribution of~$\bm{\theta}$, yielding a posterior-aware likelihood that accounts for map parameter uncertainty in particle weighting.

\subsection{Conditional Map Estimation}
\label{sec:update}

This subsection develops the first factor of Equation~\eqref{eq:rbpf}, the map posterior $P\bigl(\bm{\theta}\mid\bm{x}[1:i],\tilde{\bm{z}}[1:i]\bigr)$ conditioned on a single particle's trajectory. Each particle maintains its own such posterior, updated independently at every time step from the radar measurements of Section~\ref{sec:Processing}. For readability, the particle index $k$ is omitted in this subsection.

\textsc{Hilbert-GP} was originally developed for GP regression, where the observations are continuous and follow Gaussians, so that the posterior over $\bm{\theta}$ is itself Gaussian and available in closed form. Occupancy mapping breaks this assumption: the labels are binary ($\pm 1$), and the corresponding Bernoulli likelihood in Equation~\eqref{eq:Bernoulli} is non-Gaussian and non-conjugate to the Gaussian prior on $\bm{\theta}$, so the exact posterior is no longer closed-form. We therefore consider two approximations that restore a tractable Gaussian posterior over $\bm{\theta}$: (i)~a Kalman filter-based estimator, which treats the binary labels as Gaussian observations so that the linear-Gaussian update applies, and (ii)~a recursive Laplace approximation, which fits a Gaussian at the mode of the non-Gaussian posterior. Considering the computational complexity, we adopt the Kalman filter-based approach throughout this work and justify this choice at the end of this subsection.

\subsubsection{Kalman Filter-Based Map Update}
\label{sec:kf_update}

Occupancy value estimation can be cast as Probabilistic Least-Squares Classification (PLSC)~\cite{rasmussen2006gaussian}, which treats discrete labels as Gaussian-distributed observations. Building on the \textsc{Hilbert-GP} formulation, we realize a sequential implementation of PLSC in a Kalman filter scheme.

When a new measurement $\tilde{\bm{z}}[i]$ arrives at time step~$i$, the posterior is updated recursively as
\begin{equation}
  \label{eqUpdateMap}
  \begin{aligned}
    \hat{\bm{\theta}}[i]
      &= \hat{\bm{\theta}}[i-1]
         + \bm{K}[i]
           \bigl(\tilde{\bm{z}}[i]-\hat{\bm{z}}[i]\bigr)\\[4pt]
    \operatorname{cov}\!\bigl(\hat{\bm{\theta}}[i]\bigr)
      &= \operatorname{cov}\!\bigl(\hat{\bm{\theta}}[i-1]\bigr)
         - \bm{K}[i]\,
           \operatorname{cov}\!\bigl(\tilde{\bm{z}}[i]\bigr)\,
           \bm{K}[i]^{\top}
  \end{aligned}
\end{equation}
where $[\bm{p}_n[i]]_{n=1}^{N}$ collects the $N$ sampled positions associated with the current measurements. The predicted measurement $\hat{\bm{z}}[i]$, conditioned on the prior estimate~$\hat{\bm{\theta}}[i-1]$, is given by
\begin{equation}
  \label{eq:predictedZ}
  \hat{\bm{z}}[i]
  = \bm{H}[i]^{\top}\,\hat{\bm{\theta}}[i-1]
\end{equation}
with the $M\times N$ design matrix
\begin{equation}
  \label{eq:basismatrix}
  \begin{aligned}
    \bm{H}[i]
    &= \bm{\Phi}\!\left([\bm{p}_n[i]]_{n=1}^{N}\right)\\
    &= \begin{bmatrix}
        \bm{\Phi}(\bm{p}_1[i]) &
        \bm{\Phi}(\bm{p}_2[i]) &
        \cdots &
        \bm{\Phi}(\bm{p}_N[i])
      \end{bmatrix}
  \end{aligned}
\end{equation}
where $M$ is the number of basis functions used for map representation and $N$ is the number of measurements per scan. For brevity, we denote $\bm{\Phi}_n := \bm{\Phi}(\bm{p}_n[i])$ as the feature vector at the $n$-th sample, so that $\bm{H}[i] = [\bm{\Phi}_1, \bm{\Phi}_2, \ldots, \bm{\Phi}_N]$. The innovation covariance is
\begin{equation}
  \label{eq:covz}
  \operatorname{cov}\!\bigl(\tilde{\bm{z}}[i]\bigr)
  = \bm{H}[i]^{\top}\,
    \operatorname{cov}\!\bigl(\hat{\bm{\theta}}[i-1]\bigr)\,
    \bm{H}[i]
    + \sigma_n^2\bm{I}_N
\end{equation}
where $\sigma_n^2$ is the measurement-noise variance, characterizing the deviation between observations and the model. Together with the length-scale~$\ell$ and the signal variance~$\sigma_f^2$, $\sigma_n^2$ forms the set of three model hyperparameters. Finally, the Kalman gain is
\begin{equation}
  \label{eq:gain_matrix}
  \bm{K}[i]
  = \operatorname{cov}\!\bigl(\hat{\bm{\theta}}[i-1]\bigr)\,
    \bm{H}[i]\,
    \operatorname{cov}\!\bigl(\tilde{\bm{z}}[i]\bigr)^{-1}
\end{equation}

\subsubsection{Comparison with Laplace Approximation}
\label{sec:laplace_compare}

An alternative is the recursive Laplace approximation, which iteratively finds the mode of the posterior distribution; its full derivation is given in the Appendix. As analyzed in Table~\ref{tab:SequentialUpdate}, the Kalman filter-based update has overall complexity $\mathcal{O}(MN^2 + M^2N + N^3)$, whereas the recursive Laplace approximation includes an $\mathcal{O}(M^3)$ Newton step and typically requires several iterations to converge. Because $M>N$ in our typical implementation, i.e., the number of basis functions exceeds the number of measurements per scan, the Kalman filter-based approach is more suitable for online sequential mapping. We therefore adopt it for sequential map update throughout this work.

\subsection{Pose Estimation}
\label{sec:pf}

We now develop the second factor of Equation~\eqref{eq:rbpf}, the pose posterior $P\bigl(\bm{x}[1:i]\mid\tilde{\bm{z}}[1:i],\bm{u}[1:i]\bigr)$, which the RBPF represents with a weighted set of particles. Each particle is first propagated according to the motion model, then weighted by how well its pose hypothesis explains the radar scan under its conditional map posterior, and finally subjected to selection and resampling.

\subsubsection{Particle Representation and Propagation}
\label{sec:propagation}

Each particle in RICH-SLAM maintains a trajectory history, a particle-conditioned map posterior, and an importance weight. Following the Hilbert-space formulation in Equation~\eqref{eq:Latent}, the basis functions $\bm{\Phi}(\cdot)$ are determined by the kernel hyperparameters and the domain $\Omega$, and are therefore shared across all particles. The particle set at time step~$i$ is represented as
\begin{equation}
  \bm{\Xi}[i]
  =
  \left\{
  \left(
  \bm{x}_k[1:i],
  \hat{\bm{\theta}}_k[i],
  \operatorname{cov}(\hat{\bm{\theta}}_k[i]),
  w_k[i]
  \right)
  \right\}_{k=1}^{N_p}
\end{equation}
in which $N_p$ is the number of particles, $\bm{x}_k[1:i]$ is the trajectory history of the $k$-th particle, $\hat{\bm{\theta}}_k[i]$ and $\operatorname{cov}(\hat{\bm{\theta}}_k[i])$ define its map posterior, and $w_k[i]$ is its importance weight.

As described in Section~\ref{sec:Processing}, we adopt a general motion model that accommodates arbitrary odometry sources. Each particle is propagated according to
\begin{equation}
  \label{eqPropagation}
  \bm{x}_k[i]
    = f\!\left(\bm{x}_k[i-1],\,\bm{u}[i],\,\bm{v}_k[i]\right)
\end{equation}
where $\bm{u}[i]$ is the control input derived from odometry measurements and $\bm{v}_k[i]$ is the process noise of the $k$-th particle. This formulation supports nonlinear state transitions (motion models)~$f(\cdot)$ and non-Gaussian noise distributions~$P(\bm{v}[i])$, thereby encompassing commonly used motion models in SLAM. In practice, we adopt the motion model as the proposal distribution, i.e.,
$q(\bm{x}[i] \mid \bm{x}_k[i-1], \bm{u}[i])
=
P(\bm{x}[i] \mid \bm{x}_k[i-1], \bm{u}[i])$
following classical RBPF implementations. The covariance of $\bm{v}_k[i]$ controls the spread of the particle proposal and is kept fixed across all sequences in our experiments.

\subsubsection{Posterior-aware Weight Calculation}
\label{sec:uncertainty}

The weight of each particle quantifies how well its current pose hypothesis explains the observed measurements under its particle-conditioned map posterior. For the $k$-th particle, let $\mathcal{D}_k[i-1]$ denote the trajectory history and map posterior before incorporating the current radar scan. The weight is updated as
\begin{equation}
  \label{eq:weighting}
  w_k[i]
  =
  w_k[i-1]\,
  Pr\!\left(
  \tilde{\bm{z}}[i]
  \mid
  \bm{x}_k[i],\,\mathcal{D}_k[i-1]
  \right)
\end{equation}

Assuming conditional independence across the $N$ sampled points, the joint likelihood factorizes as
\begin{equation}
\small
  Pr\!\left(
  \tilde{\bm{z}}[i]
  \mid
  \bm{x}_k[i],\,\mathcal{D}_k[i-1]
  \right)
  \approx
  \prod_{n=1}^{N}
  Pr\!\left(
  (\tilde{\bm{z}}[i])_n
  \mid
  \bm{x}_k[i],\,\mathcal{D}_k[i-1]
  \right)
\end{equation}

For the $n$-th sample, the likelihood is obtained by marginalizing over the latent occupancy value
\begin{equation}
\small
  \label{eqPointLikelihood}
  \begin{aligned}
    &Pr\!\left(
      (\tilde{\bm{z}}[i])_n
      \mid
      \bm{x}_k[i],\,\mathcal{D}_k[i-1]
      \right)\\
    &=\int
       Pr\!\left((\tilde{\bm{z}}[i])_n
         \mid(\bm{z}[i])_n\right)\,
       P\!\left((\bm{z}[i])_n
         \mid\bm{x}_k[i],\,\mathcal{D}_k[i-1]\right)
       \mathrm{d}(\bm{z}[i])_n
  \end{aligned}
\end{equation}
where $Pr\!\left((\tilde{\bm{z}}[i])_n \mid(\bm{z}[i])_n\right)$ is the measurement model defined in Equation~\eqref{eq:Bernoulli}. To account for uncertainty in~$\bm{\theta}$, the predictive distribution of the latent value is obtained by marginalizing over the particle-conditioned map posterior:
\begin{equation}
  \label{eq:predictDistribution}
  P\!\left(
  (\bm{z}[i])_n
  \mid
  \bm{x}_k[i],\,\mathcal{D}_k[i-1]
  \right)
  =
  \mathcal{N}\!\left(
  \hat{z}_{k,n}[i],
  s^2_{k,n}[i]
  \right)
\end{equation}
where
\begin{equation}
  \hat{z}_{k,n}[i]
  =
  \bm{\Phi}(\bm{p}_{k,n}[i])^\top
  \hat{\bm{\theta}}_k[i-1]
\end{equation}
and
\begin{equation}
  s^2_{k,n}[i]
  =
  \bm{\Phi}(\bm{p}_{k,n}[i])^\top
  \operatorname{cov}\!\left(
  \hat{\bm{\theta}}_k[i-1]\right)
  \bm{\Phi}(\bm{p}_{k,n}[i])
\end{equation}
in which $\bm{p}_{k,n}[i]=\mathcal{T}(\bm{x}_k[i],\Delta\bm{p}_n^b)$ is the $n$-th sampled point transformed by the $k$-th particle pose.

Combining Equations~\eqref{eq:Bernoulli} and~\eqref{eq:predictDistribution} and exploiting the resemblance between the sigmoid and probit functions~\cite{bishop2006pattern}, we define the posterior-predictive occupancy probability queried by the $k$-th particle as
\begin{equation}
  \label{eq:posterior_predictive_occupancy}
  o_k(\bm{p})
  =
  \varsigma\!\left(
  \dfrac{
  \bm{\Phi}(\bm{p})^\top \hat{\bm{\theta}}_k[i-1]}
  {\sqrt{
  1+\frac{\pi}{8}
  \bm{\Phi}(\bm{p})^\top
  \operatorname{cov}(\hat{\bm{\theta}}_k[i-1])
  \bm{\Phi}(\bm{p})}}
  \right)
\end{equation}
\par This posterior-predictive form naturally accounts for map uncertainty: when the predictive variance is large, the occupancy probability is driven toward~0.5, reflecting low confidence; when the variance is small, the probability becomes more decisive.

Sparse radar endpoints, however, should not be treated as exact geometric hits because radar returns are affected by range-angle noise, specular reflection, discretization, and small pose errors. Therefore, for positive endpoint samples, we evaluate the occupancy likelihood over a local neighborhood rather than only at the exact endpoint location. For a query point $\bm{p}$, we define
\begin{equation}
  \label{eq:endpoint_tolerance}
  \tilde{o}_k(\bm{p})
  =
  \max_{\|\bm{q}-\bm{p}\| \le r_e}
  o_k(\bm{q})
\end{equation}
The per-sample likelihood used for particle weighting is then
\begin{equation}
  \label{eq:probit_approx}
  \begin{aligned}
  &Pr\!\left((\tilde{\bm{z}}[i])_n
  \mid
  \bm{x}_k[i],\,\mathcal{D}_k[i-1]\right) \\
  &\quad \approx
  \begin{cases}
    \tilde{o}_k(\bm{p}_{k,n}[i]),
    & (\tilde{\bm{z}}[i])_n=1,\\[3pt]
    1-o_k(\bm{p}_{k,n}[i]),
    & (\tilde{\bm{z}}[i])_n=-1.
  \end{cases}
  \end{aligned}
\end{equation}
when $(\tilde{\bm{z}}[i])_n=1$, the sampled point $\bm{p}_{k,n}[i]$ corresponds to the radar endpoint. In the experiments, we set $r_e=\SI{0.25}{\meter}$. This endpoint-tolerant likelihood only modifies the particle weighting step and does not change the map update or the evaluation protocol. Figure~\ref{fig:BeamSampling} presents a graphical illustration of the endpoint tolerance scheme.

\subsubsection{Particle Selection and Resampling}
\label{sec:resampling}

At each time step, the particle with the largest weight is selected as the current estimate of the vehicle state and map. The weights are then normalized, and particles are resampled with probability proportional to their normalized weights~\cite{gordon1993novel}, ensuring that high-likelihood particles propagate while low-likelihood ones are discarded during this process.

To prevent particle depletion caused by resampling, resampling is performed only when the effective sample size~\cite{barkby2012bathymetric}
\begin{equation}
N_{\text{eff}} =\frac{\bigl(\sum_{k=1}^{N_p} w_k[i]\bigr)^2}{\sum_{k=1}^{N_p} \bigl(w_k[i]\bigr)^2}
\end{equation}
falls below half the total number of particles~\cite{liu1996metropolized}. This strategy preserves particle diversity and prevents premature convergence on suboptimal states.

\subsection{RICH-SLAM Pipeline}
\label{sec:pipeline}

Finally, Algorithm~\ref{alg:particle_filter} assembles the two factors of Equation~\eqref{eq:rbpf} into a single recursive loop. At each time step~$i$, every particle is first propagated (Section~\ref{sec:propagation}) and weighted under its prior map posterior $\mathcal{D}_k[i-1]$ (Section~\ref{sec:uncertainty}). The current radar measurements are then incorporated into that particle's map posterior via the Kalman update of Section~\ref{sec:update}, yielding $\hat{\bm{\theta}}_k[i]$ and $\operatorname{cov}(\hat{\bm{\theta}}_k[i])$, which serve as the prior at the next time step. Finally, particle selection and resampling (Section~\ref{sec:resampling}) conclude the iteration. Because the weight in Equation~\eqref{eq:weighting} is evaluated against the prior map posterior, the map update is applied \emph{after} weighting.

\begin{algorithm}[t]
\caption{RICH-SLAM Pipeline}
\label{alg:particle_filter}
\begin{algorithmic}[1]
\State Initialize $\bm{x}_{k}[0]$, $\hat{\bm{\theta}}_{k}[0]$,
       $\operatorname{cov}(\hat{\bm{\theta}}_{k}[0])$, and $w_k[0]=1/N_p$
       for $k = 1,\dots,N_p$
\For{$i = 1, \dots, T_{\max}$}
    \For{$k = 1, \dots, N_p$}
        \State Draw sample
               $\bm{x}_{k}[i]
                \sim q(\bm{x}[i]\mid\bm{x}_{k}[i-1],\bm{u}[i])$
        \State Compute weight $w_k[i]$
               using Equation~\eqref{eq:weighting}
        \State Update map posterior
               using Equation~\eqref{eqUpdateMap}
    \EndFor
    \State $k^\star \gets \arg\max_k\, w_k[i]$
    \State Output $\bm{x}_{k^\star}[i]$ and
           $\hat{\bm{\theta}}_{k^\star}[i]$ as current estimates
    \If{$N_\mathrm{eff} < 0.5\,N_p$}
        \State Resample $N_p$ particles
        \State $w_k[i] \gets 1/N_p$ for all $k$
    \EndIf
\EndFor
\end{algorithmic}
\end{algorithm}

\section{Experimental Results}
\label{sec:exp}

The experiments are designed to address the following questions:
\begin{itemize}
    \item[\textbf{Q1.}] What compactness and accuracy trade-offs does \textsc{Hilbert-GP} provide relative to feature-based GP approximations?
    \item[\textbf{Q2.}] Which kernel hyperparameters affect mapping quality, and what properties does the continuous representation provide?
    \item[\textbf{Q3.}] Can RICH-SLAM improve localization while building continuous maps online?
    \item[\textbf{Q4.}] Can the resulting continuous maps support downstream path planning?
\end{itemize}

We first describe the experimental setup in Section~\ref{sec:setup}, then address
\textbf{Q1} and \textbf{Q2} through mapping evaluation in
Section~\ref{sec:mappingEval}, \textbf{Q3} through online mapping visualization
and localization evaluation in Sections~\ref{sec:mappingEval} and
\ref{sec:localizationEval}, and \textbf{Q4} through a path planning demonstration
in Section~\ref{sec:planning}. Section~\ref{sec:discussion} reports our findings
with a detailed discussion.

\subsection{Set Up}
\label{sec:setup}

\subsubsection{Evaluation Protocol}

\begin{figure}[!t]
    \centering
    \includegraphics[width=0.99\linewidth]{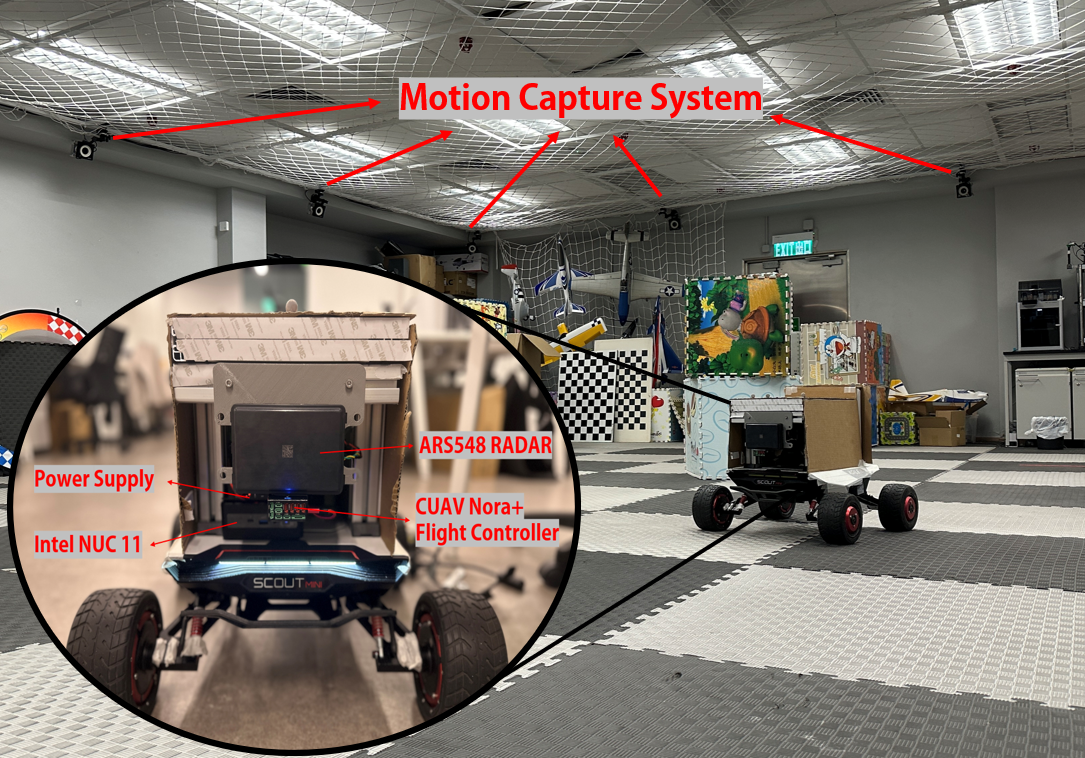}
    \caption{Experimental setup of self-collected dataset. The vehicle is equipped with a radar sensor and an IMU. Experimental validation was performed in a controlled indoor environment using a motion capture system to provide ground-truth poses.}
    \label{fig:selfdataset}
\end{figure}

We evaluate RICH-SLAM on two datasets that employ different radar sensors and span diverse environments. The first is our self-collected dataset, acquired from a mobile sensing platform (Figure~\ref{fig:selfdataset}) equipped with a Continental ARS548RDI 4D FMCW radar and a Bosch BMI088 IMU. A motion capture system provides millimeter-level ground truth accuracy. We collect three test sequences of increasing complexity: Sequence~1 (low-speed rectangular trajectory), Sequence~2 (higher-speed rectangular trajectory), and Sequence~3 (high-speed complex maneuvers).

\begin{figure}[t]
    \centering
    \subfloat[Platform]{
        \label{figPlatform}
        \includegraphics[width=0.48\linewidth]{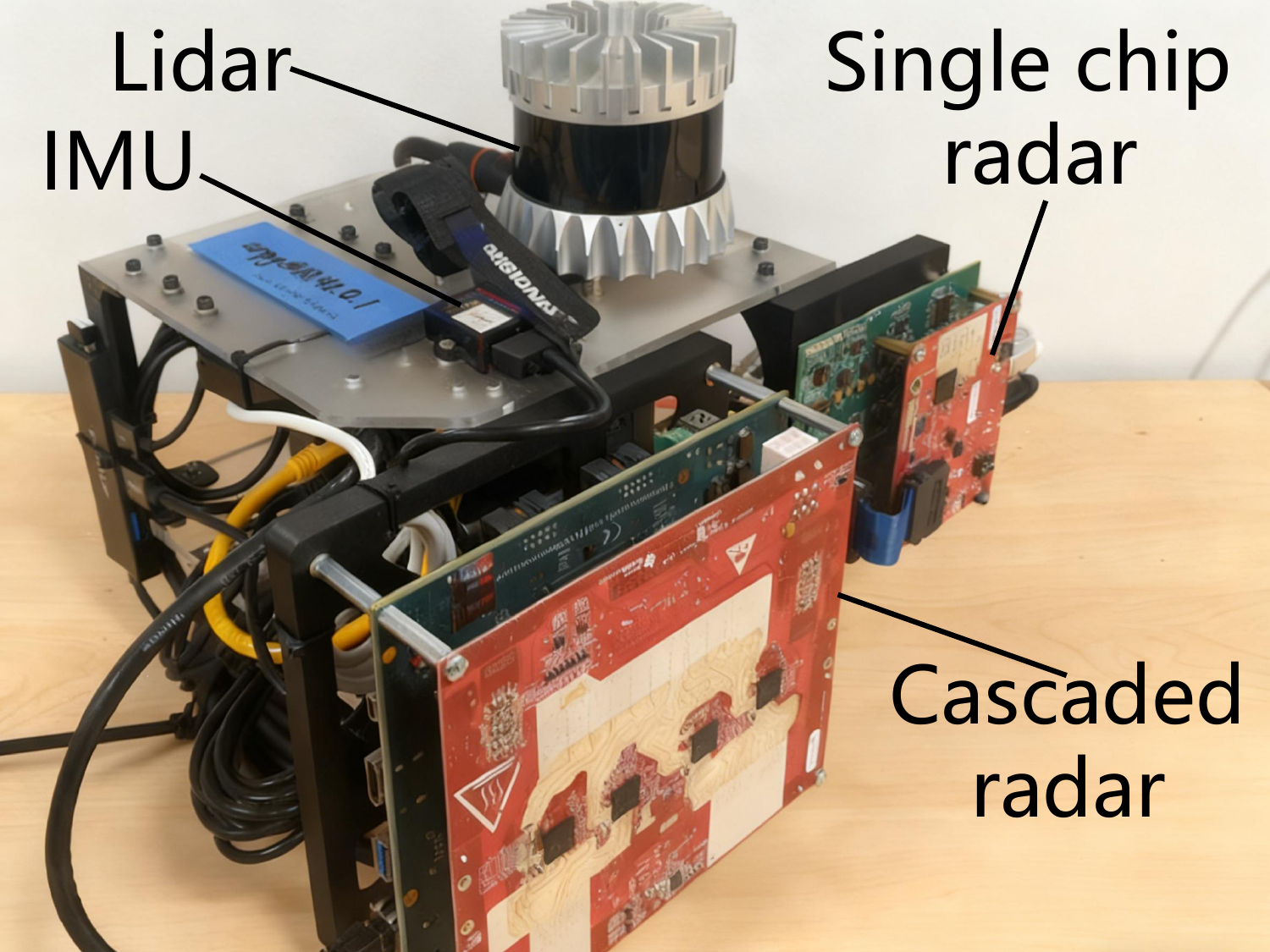}
    }
    \subfloat[ColoRadar 1]{
        \label{figColoRadar1}
        \includegraphics[width=0.48\linewidth]{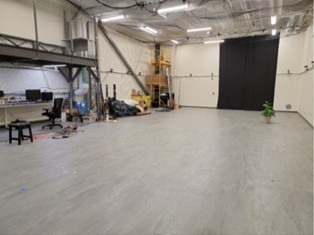}}
        \\
    \subfloat[ColoRadar 2]{
        \label{figColoRadar2}
        \includegraphics[width=0.48\linewidth]{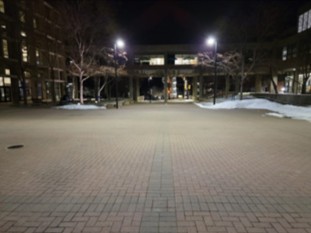}
    }
    \subfloat[ColoRadar 3 and 6]{
        \label{figColoRadar3}
        \includegraphics[width=0.48\linewidth]{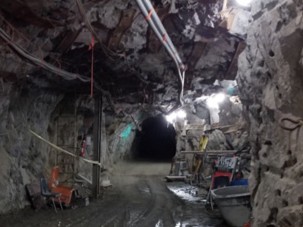}
    }
    \\
    \subfloat[ColoRadar 4]{
        \label{figColoRadar4}
        \includegraphics[width=0.48\linewidth]{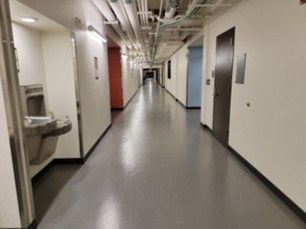}
    }
    \subfloat[ColoRadar 5]{
        \label{figColoRadar5}
        \includegraphics[width=0.48\linewidth]{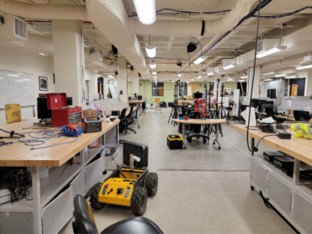}
    }
    \caption{Experimental setup of ColoRadar datasets (Figures in \cite{kramer2022ColoRadar}). For certain datasets, ground truth trajectories are acquired through LiDAR-IMU fusion, ensuring accurate reference data for evaluation.}
    \label{fig:ColoRadar}
\end{figure}

The second is the ColoRadar dataset~\cite{kramer2022ColoRadar}, with a single-chip Texas Instruments radar sensor that produces sparse point clouds (Figure~\ref{fig:ColoRadar}). We select six sequences spanning three environment types: indoor (ColoRadar~1, ColoRadar~4, ColoRadar~5), outdoor (ColoRadar~2), and subterranean (ColoRadar~3 and ColoRadar~6) \footnote{ColoRadar 1-6 corresponds to ``2\_24\_2021\_aspen\_run0'', ``2\_28\_2021\_outdoor\_run0'', ``2\_23\_2021\_edgar\_classroom\_run0'', ``12\_21\_2020\_ec\_hallways\_run0'', ``12\_21\_2020\_arpg\_lab\_run0'' and ``2\_23\_2021\_edgar\_classroom\_run3'', respectively}. Overall, the two datasets cover two distinct radar modalities and a range of environmental settings, thus enabling a comprehensive assessment of the proposed RICH-SLAM system.

\subsubsection{Implementation Details}

For odometry, we employ the recent radar-inertial method of~\cite{huang2024less} to estimate ego motion. At the back end, the particle filter is initialized with $N_p=200$ particles. Although the two datasets use the same \textsc{Hilbert-GP} mapping framework, their radar return statistics are different due to the use of different radar sensors. Therefore, we use dataset-level mapping parameters, and the key parameters are summarized in Table~\ref{tab:rich_slam_mapping_params}. The length scale and basis size are kept identical across datasets to preserve the same GP smoothness prior and computational budget. The Hilbert-domain half-widths $(L_x,L_y)$ are sequence-specific because they define the finite spatial support of the map and therefore scale with the real-world environment. For all sequences, $(L_x,L_y)$ are determined using the same trajectory-envelope rule with fixed padding, rather than tuned from the mapping or localization results. In contrast, the GP signal and noise variances are determined at the dataset level: the TI single-chip radar in ColoRadar produces sparser and noisier raw data than the ARS548RDI radar, thus requiring a larger signal variance to maintain occupancy contrast and a larger noise variance to avoid overfitting spurious radar returns. Except for the environment-dependent Hilbert domain, the parameters are kept unchanged for all sequences within each dataset.

\begin{figure*}[t]
    \centering
    \includegraphics[width=0.99\linewidth]{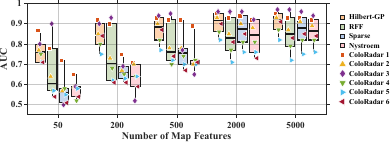}
    \caption{The area under the receiver operating characteristic curve (AUC) values versus different numbers of map features. For \textsc{Hilbert-GP}, this number denotes the retained Hilbert-space basis functions; for \textsc{RFF}, \textsc{Sparse}, and \textsc{Nystr\"om}, it denotes the feature components used by the corresponding map representation. \textsc{Hilbert-GP} achieves consistently high AUC with compact map representations, while the SGD-based baselines exhibit larger variance across datasets.} 
    \label{fig:ParamSize}
\end{figure*}

\begin{figure*}[t]
    \centering
    \includegraphics[width=0.99\linewidth]{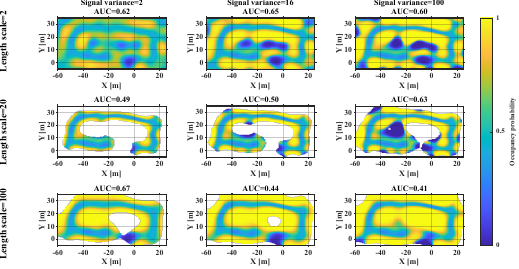}
    \caption{Mapping results under different length scales and signal variances. White regions correspond to locations not included in the queried map domain, rather than free space. The length scale controls the spatial smoothness of the occupancy field: a small length scale preserves local variation but can produce fragmented structures, whereas a large length scale over-smooths the map and spreads high occupancy probability over broad regions. The signal variance controls the amplitude of the latent occupancy function and affects the contrast and confidence of the predicted occupancy map. AUC values are reported for reference, while parameter selection also considers qualitative map continuity.}
    \label{fig:Hyperparam}
\end{figure*}

\begin{table}[t]
\centering
\caption{Mapping hyperparameters for RICH-SLAM. }
\label{tab:rich_slam_mapping_params}
\footnotesize
\setlength{\tabcolsep}{4pt}
\renewcommand{\arraystretch}{1.15}
\begin{tabular}{lcc}
\hline\hline
\textbf{Parameter} & \textbf{Self-collected} & \textbf{ColoRadar} \\
\hline
Sequences & Seq.~1--3 & CR~1--6 \\
Radar sensors & ARS548RDI & TI single-chip radar \\
Length scale $\ell$ & 20 & 20 \\
Signal variance $\sigma_f^2$ & 0.7 & 16 \\
Measurement variance $\sigma_n^2$ & 0.1 & 0.36 \\
Basis functions $M$ & 256 & 256 \\
Maximum radar range [m] & 5 & 5 \\
Hilbert domain $(L_x,L_y)$ & sequence-specific & sequence-specific \\
Ray sampling step [m] & 1 & 1 \\
\hline\hline
\end{tabular}
\end{table}


\subsection{Mapping Performance Evaluation}
\label{sec:mappingEval}

In this section, we evaluate the mapping module of RICH-SLAM from three aspects: comparison with GP approximation baselines, hyperparameter sensitivity, and incremental mapping together with the properties of the resulting representation: resolution-independent querying and predictive uncertainty.

\subsubsection{Comparison with GP Approximation Baselines (\textbf{Q1})}

We first evaluate occupancy prediction accuracy with the area under the receiver operating characteristic curve (AUC)~\cite{ramos2016hilbert}. Ground-truth poses are used to align radar observations for the quantitative AUC evaluation, so that mapping accuracy can be assessed independently of pose-estimation errors. For this quantitative AUC comparison, 80\% of each dataset serves as the training set and the remaining 20\% is held out for testing. In addition to AUC, we report the number of map features used by each map representation, which serves as a proxy for map compactness and memory cost.

We compare \textsc{Hilbert-GP} with three feature-based GP approximations from~\cite{ramos2016hilbert}: Random Fourier Features (\textsc{RFF}), Sparse Random Features (\textsc{Sparse}), and Nystr\"om Features (\textsc{Nystr\"om}). The SGD update is used for the three baselines following the original Hilbert-map formulation, which treats the feature weights as deterministic parameters of a logistic occupancy classifier. In contrast, \textsc{Hilbert-GP} maintains a posterior distribution over the map weights and therefore admits a Kalman-filter update. Regarding the number of map features, it is the number of Hilbert-space basis functions $M$ for \textsc{Hilbert-GP}; for \textsc{RFF}, \textsc{Sparse}, and \textsc{Nystr\"om}, it is the number of random or data-dependent features used by the corresponding logistic occupancy model. Thus,the number of map features serves as a parameter cost to measure the map compactness.

As shown in Figure~\ref{fig:ParamSize}, \textsc{Hilbert-GP} achieves a more compact map representation than \textsc{Sparse} and \textsc{Nystr\"om}. Although \textsc{RFF} attains higher AUC on certain individual datasets, \textsc{Hilbert-GP} exhibits better stability across all datasets. We attribute this stability to its uncertainty estimates, which effectively guide incremental map updates. More specifically, the Kalman filter update (Equation~\eqref{eqUpdateMap}) adjusts each weight $\theta_j$ by an amount proportional to the Kalman gain, which itself depends on the current posterior covariance. When a region has been well observed, the covariance is small and new measurements produce only minor corrections, preventing the map from oscillating. In contrast, the SGD-based methods lack an explicit uncertainty model and apply a fixed or decaying learning rate uniformly, which can cause the map to fluctuate when revisiting previously mapped areas. This distinction is most apparent on the ColoRadar sequences, where the robot revisits corridors multiple times: \textsc{Hilbert-GP} maintains a consistently high AUC across different numbers of map features (Figure~\ref{fig:ParamSize}), whereas the SGD baselines exhibit larger variance across runs.

\subsubsection{Hyperparameter Sensitivity (\textbf{Q2})}

We further analyze the hyperparameter sensitivity of \textsc{Hilbert-GP}. Figure~\ref{fig:Hyperparam} reports mapping results under different length scales $\ell$ and signal variances $\sigma_f^2$, with the measurement variance fixed at $\sigma_n^2=0.36$. The results show that the length scale has a strong influence on the spatial structure of the reconstructed occupancy map. When $\ell$ is small ($\ell=2$), the map preserves more local variation, but it also produces fragmented occupancy patterns and less spatially coherent structures. When $\ell$ is too large ($\ell=100$), the kernel over-smooths the occupancy field, producing broad high-confidence regions and erasing local free-space and occupied-space details. The intermediate value $\ell=20$ provides a more balanced map representation: the main corridor-like structures remain continuous, while local details are not over-smoothed as aggressively as in the large-length-scale case.

The signal variance $\sigma_f^2$ controls the amplitude of the latent occupancy function. A small signal variance tends to suppress occupancy contrast, whereas an excessively large signal variance can make the map overly confident in sparsely observed regions. Although the highest AUC in Figure~\ref{fig:Hyperparam} is not always obtained by the selected dataset-level parameters, AUC is not the only criterion for choosing online mapping parameters. AUC measures point-wise occupancy classification accuracy on held-out samples, whereas RICH-SLAM also requires a continuous map with reasonable spatial smoothness, structural consistency, and stable uncertainty behavior for localization and planning. From the qualitative maps, $\ell=20$ and $\sigma_f^2=16$ provide a practical compromise between occupancy contrast and spatial continuity. These observations motivate the dataset-level GP hyperparameters in Table~\ref{tab:rich_slam_mapping_params}, where the length scale is kept fixed to preserve the same spatial representation, while the signal and measurement variances are calibrated according to the radar return statistics of each dataset.

\subsubsection{Visualization of Incremental Mapping (\textbf{Q3})}

\begin{figure*}[!t]
    \centering
    \subfloat[Sequence 3]{
        \label{fig:MappingSeq3}
        \includegraphics[width=0.95\linewidth]{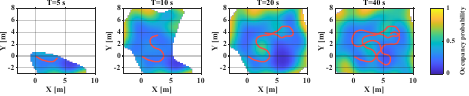}
    }
    \\
    \subfloat[ColoRadar 2]{
        \label{fig:MappingColoradar2}
        \includegraphics[width=0.95\linewidth]{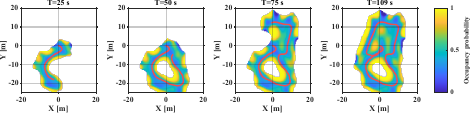}
    }
    \\
    \subfloat[ColoRadar 3]{
        \label{fig:MappingColoradar3}
        \includegraphics[width=0.95\linewidth]{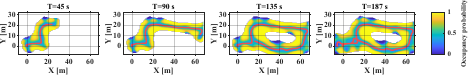}
    }
    \\
    \subfloat[ColoRadar 4]{
        \label{fig:MappingColoradar4}
        \includegraphics[width=0.95\linewidth]{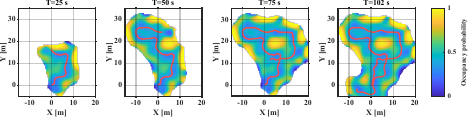}
    }
    \caption{Visualization of online mapping with RICH-SLAM on Sequence~3,
    ColoRadar~2, ColoRadar~3, and ColoRadar~4. Occupancy probabilities range
    from 0 (lowest likelihood of occupancy) to 1 (highest likelihood of
    occupancy), represented by a color gradient from blue to yellow. The red
    curves show the trajectories accumulated up to each timestamp. }
    \label{fig:Mapping}
\end{figure*}

Figure~\ref{fig:Mapping} presents the incremental mapping process of RICH-SLAM with \textsc{Hilbert-GP} across four representative environments. In all cases, the map is updated online as new radar beams arrive, without reprocessing previous observations. Unlike the evaluations above, these qualitative online mapping results are generated from the RICH-SLAM estimated trajectories.

Across the four sequences in Figure~\ref{fig:Mapping}, the map expands consistently as new radar beams are incorporated online. In the self-collected Sequence~3 (Figure~\ref{fig:MappingSeq3}), the explored area grows from a local region at $T=\SI{5}{\second}$ to a larger loop-shaped map at $T=\SI{40}{\second}$ despite faster and more complex motion. ColoRadar~2 (Figure~\ref{fig:MappingColoradar2}) illustrates an outdoor case with sparse returns and larger open regions, where the map gradually expands from $T=\SI{25}{\second}$ to $T=\SI{109}{\second}$ while unobserved regions remain close to the prior. ColoRadar~3 and ColoRadar~4 (Figures~\ref{fig:MappingColoradar3} and~\ref{fig:MappingColoradar4}) further show tunnel-like and hallway-like environments, where the explored region extends along elongated corridor structures and repeated observations sharpen the occupied boundaries. These results demonstrate that RICH-SLAM can incrementally construct continuous occupancy maps across compact indoor, outdoor, and corridor-like environments.

\begin{figure*}[!t]
    \centering
    \includegraphics[width=0.95\linewidth]{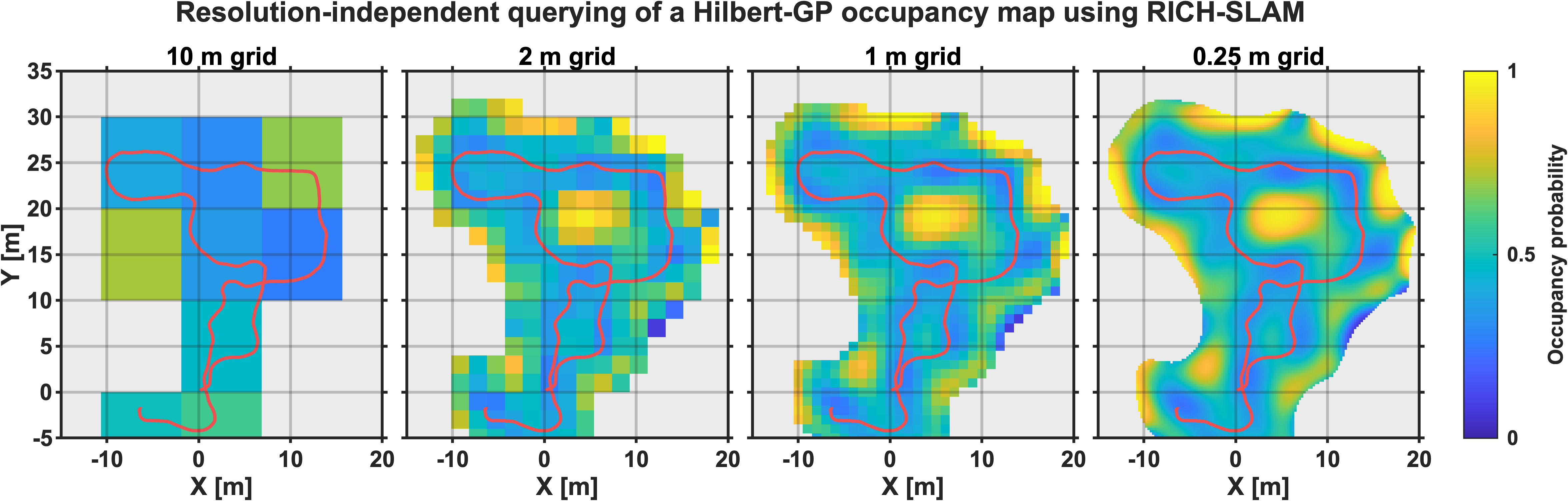}
    \caption{Resolution-independent querying of a \textsc{Hilbert-GP} occupancy map using RICH-SLAM on ColoRadar~4. The map is learned once using the same $M=256$ Hilbert basis functions and the same posterior weights. The four panels evaluate this fixed continuous occupancy function on query grids with spacings of \SI{10}{\meter}, \SI{2}{\meter}, \SI{1}{\meter}, and \SI{0.25}{\meter}, respectively. Finer query grids reveal more visual detail without retraining the map or changing the stored map representation.}
    \label{fig:resolution_independent_query}
\end{figure*}

In addition to online map growth, the \textsc{Hilbert-GP} representation provides a resolution-independent query interface. Unlike a discrete occupancy grid whose cell size is fixed at map construction time, the RICH-SLAM map stores a continuous latent occupancy function parameterized by Hilbert basis weights. Once the map is learned, occupancy probabilities can be evaluated at arbitrary query locations. Figure~\ref{fig:resolution_independent_query} demonstrates this property on ColoRadar~4: the same learned map is queried using grid spacings from \SI{10}{\meter} to \SI{0.25}{\meter}. Coarser grids provide compact low-resolution previews, whereas finer grids recover more detailed map structure without changing the learned model. This property is useful for post-processing, visualization, and downstream planning, where different tasks may require different spatial query resolutions.

\begin{figure*}[t]
    \centering
    \subfloat[ColoRadar 2]{
        \label{fig:UncertaintyColoradar2}
        \includegraphics[width=0.40\linewidth]{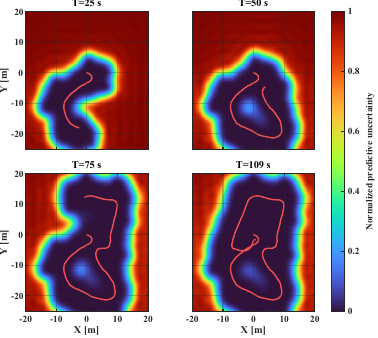}
    }
    \subfloat[Uncertainty statistics]{
        \label{fig:RichSlamUncertainty}
        \includegraphics[width=0.52\linewidth]{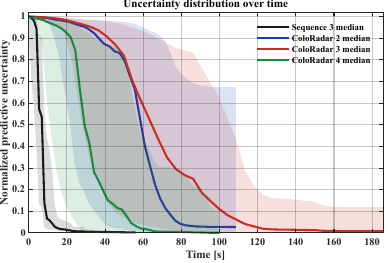}
    }

    \caption{Predictive uncertainty of the \textsc{Hilbert-GP} map. The left panel shows the normalized predictive uncertainty on ColoRadar~2 at different timestamps. Low uncertainty values appear around repeatedly observed regions, whereas high values remain in unobserved or weakly observed areas. The right panel summarizes uncertainty statistics over time on four sequences. Each color denotes one sequence, the solid curve shows the median normalized predictive uncertainty over the mapped area, and the translucent band with the same color denotes the interquartile range (25th--75th percentiles). Overall, the figure shows that predictive uncertainty decreases as radar observations accumulate.}
    \label{fig:Uncertainty}
\end{figure*}

Figure~\ref{fig:Uncertainty} further visualizes the predictive uncertainty produced by the \textsc{Hilbert-GP} map using the ColoRadar~2. The uncertainty is initially high over most of the domain and gradually decreases along the explored trajectory as more radar observations are incorporated. Regions that are repeatedly observed become more confident, whereas areas outside the radar coverage remain close to the prior. The temporal statistics show the same trend across Sequence~3, ColoRadar~2, ColoRadar~3, and ColoRadar~4: the median uncertainty decreases as mapping progresses, confirming that the proposed incremental update not only estimates occupancy but also maintains a meaningful uncertainty field.

\begin{figure*}[t]
    \centering
    \includegraphics[width=0.95\linewidth]{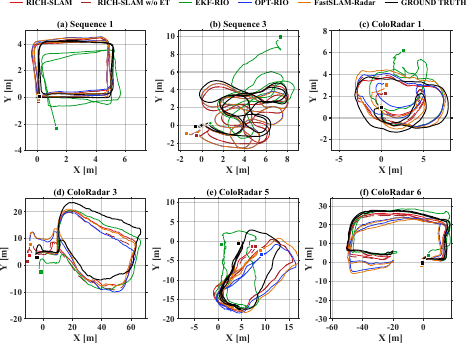}
    \caption{Comparisons of the localization performance on six representative sequences. Circles and squares represent the starting and ending points of the trajectories, respectively.}
    \label{fig:Traj}
\end{figure*}

\begin{table*}[t]
\centering
\caption{Comparison of localization performance. The reported RICH-SLAM result uses endpoint-tolerance while RICH-SLAM w/o ET disables this endpoint tolerance. Each entry reports horizontal position RMSE [m] / yaw RMSE [deg].}
\label{tab:localization_common_gt_first_100_frame_se2_two_group_rich_wide}
\scriptsize
\setlength{\tabcolsep}{2.4pt}
\begin{tabular}{lccccccccc}
\hline
\hline
Method & Seq. 1 & Seq. 2 & Seq. 3 & CR 1 & CR 2 & CR 3 & CR 4 & CR 5 & CR 6 \\ Length / Duration
 & 34.93 m, 112 s & 38.05 m, 74 s & 66.38 m, 56 s & 45.07 m, 86 s & 114.64 m, 109 s & 185.73 m, 187 s & 109.88 m, 100 s & 96.71 m, 119 s & 301.55 m, 275 s \\
\midrule
EKF-RIO & 
1.01 / 7.28 & 
2.17 / 4.03 & 
4.70 / 104.11 & 
3.26 / 79.53 & 
\textbf{1.60} / 91.98 & 
\textbf{3.72} / 97.11 & 
\textbf{0.89} / 96.45 & 
\underline{2.76} / 79.15 & 
\underline{3.26} / 86.98 \\

OPT-RIO & 
\underline{0.52} / \textbf{1.97} & 
\textbf{0.49} / \textbf{1.28} & 
\underline{1.46} / \underline{14.73} & 
1.28 / \underline{7.69} & 
2.78 / 25.41 & 
6.54 / \underline{15.36} & 
4.70 / \underline{16.90} & 
3.31 / 26.34 & 
5.95 / \underline{9.00} \\

FastSLAM-Radar & 
\textbf{0.50} / \underline{2.28} & 
\underline{0.50} / \underline{1.45} & 
\underline{1.46} / 14.87 & 
0.96 / 13.51 & 
\underline{2.41} / \underline{24.69} & 
5.71 / 13.04 & 
4.91 / 17.49 & 
2.97 / \underline{24.37} & 
6.64 / 10.07 \\

RICH-SLAM w/o ET & 
0.54 / 3.03 & 
0.60 / 2.70 & 
\textbf{0.94} / \textbf{7.31} & 
\textbf{0.84} / 7.07 & 
3.06 / 24.43 & 
6.29 / 12.21 & 
4.04 / 13.34 & 
2.61 / 23.81 & 
2.12 / 3.17 \\

RICH-SLAM & 
0.54 / 3.03 & 
0.60 / 2.70 & 
\textbf{0.94} / \textbf{7.31} & 
\underline{0.86} / \textbf{6.18} & 
2.68 / \textbf{22.88} & 
\underline{4.72} / \textbf{10.94} & 
\underline{3.59} / \textbf{12.61} & 
\textbf{2.55} / \textbf{23.23} & 
\textbf{1.79} / \textbf{3.06} \\
\hline
\hline
\end{tabular}
\end{table*}

\subsection{Localization Performance Evaluation}
\label{sec:localizationEval}

To answer \textbf{Q3}, we evaluate the localization accuracy of RICH-SLAM against three baselines that operate on sparse radar input: EKF-RIO~\cite{doer2020ekf}, an EKF-based radar-inertial odometry method; OPT-RIO~\cite{xu2025incorporating}, an optimization-based radar-inertial odometry method that provides a strong odometry-only reference for RICH-SLAM; and FastSLAM-Radar, a particle-filter SLAM baseline using radar sensing as input. FastSLAM-Radar and RICH-SLAM use the same RIO front end as the odometry proposal, so their comparison mainly reflects the effect of the map representation and particle-weighting likelihood.

Table~\ref{tab:localization_common_gt_first_100_frame_se2_two_group_rich_wide} reports the quantitative results on all nine sequences, with RICH-SLAM w/o ET included as a light ablation that disables endpoint tolerance. We also notice that on the three self-collected sequences, RICH-SLAM and RICH-SLAM w/o ET yield identical results. This is expected rather than coincidental: the ARS548RDI radar produces denser and less noisy returns than the TI single-chip radar, so the filtered endpoints already lie close to the underlying occupied structures. In this regime, the neighborhood maximization in Equation~\eqref{eq:endpoint_tolerance} rarely alters the relative ordering of particle weights, and in particular never changes the largest-weight particle selected as the state estimate, leaving the reported trajectories unchanged. The benefit of endpoint tolerance therefore manifests primarily on the sparser and noisier ColoRadar sequences.

Figure~\ref{fig:Traj} provides qualitative trajectory comparisons on six representative sequences. Both the table and the plotted trajectories use a common SE(2) transform estimated from the first 100 ground-truth-matched frames. Overall, RICH-SLAM achieves the best yaw RMSE on most sequences and strong position accuracy on several challenging sequences. Below we discuss the results in greater detail.


\subsubsection{Sequences Where RICH-SLAM Excels}

RICH-SLAM performs well when the radar observations provide informative geometric constraints for map-based correction. On Sequence~3, which contains faster and more complex maneuvers than the rectangular sequences, RICH-SLAM achieves better localization performance compared to others. This result suggests that the particle-filter back end is beneficial when the odometry proposal becomes uncertain and the pose posterior may deviate from a single Gaussian mode.

On ColoRadar~1, ColoRadar~5, and ColoRadar~6, RICH-SLAM also shows strong position accuracy. These sequences contain structured indoor or subterranean environments with clear boundaries, which produce consistent radar returns. The continuous occupancy map captures these structures and provides informative likelihood scores during the particle weighting step (Equation~\eqref{eq:weighting}), thereby sharpening the pose posterior. The endpoint-tolerant likelihood further makes the weighting less sensitive to small spatial offsets of sparse radar endpoints. In Figure~\ref{fig:Traj}, the RICH-SLAM trajectories remain closer to the ground truth in these representative structured environments, while the odometry-only and discrete-map baselines exhibit larger drift or less stable trajectory shapes.

\subsubsection{Sequences Where RICH-SLAM Is Less Effective}

RICH-SLAM is less effective in position accuracy on Sequence~1, Sequence~2, ColoRadar~2, ColoRadar~3, and ColoRadar~4. On Sequence~1 and Sequence~2, OPT-RIO achieves slightly lower position and yaw RMSE, suggesting that the odometry front end is already sufficiently accurate for these short rectangular trajectories. The additional map-based correction in RICH-SLAM provides limited benefit in these relatively simple cases.

On ColoRadar~2, ColoRadar~3, and ColoRadar~4, RICH-SLAM improves heading accuracy but does not achieve the lowest position RMSE. We identify two contributing factors. First, the outdoor environment in ColoRadar~2 contains larger open areas and sparse nearby reflectors, so the occupancy map provides weaker translational constraints. Second, ColoRadar~3 and ColoRadar~4 include elongated corridor-like structures, where the map likelihood is more informative for correcting orientation than for resolving translation along the corridor direction. These observations suggest that RICH-SLAM benefits most when the environment contains sufficient geometric structure to constrain both position and heading.

\subsubsection{Yaw Estimation}

Heading accuracy warrants separate attention because yaw errors compound over time and directly affect map consistency. Across the evaluated sequences, RICH-SLAM achieves better yaw RMSE on most cases, especially on the ColoRadar sequences. This trend indicates that the map-based particle weighting is effective for heading correction: even a small rotation error shifts the sampled radar points relative to the continuous occupancy map, producing a measurable change in likelihood.

The cases where RICH-SLAM does not obtain the best yaw RMSE are the shorter self-collected rectangular sequences, where the optimization-based odometry baseline is already highly accurate. In more challenging sequences with stronger drift or more complex motion, the posterior-aware map likelihood becomes more useful for correcting heading and maintaining trajectory consistency.

\subsection{Path Planning Demonstration}
\label{sec:planning}

To further address \textbf{Q4}, we conduct a path planning demonstration on a partially constructed RICH-SLAM map from ColoRadar~6. This experiment evaluates whether the continuous Hilbert map can be used as a downstream traversability representation, and whether predictive uncertainty can influence planning when the map is incomplete.

Given a start position $\bm{p}_s$ and a goal position $\bm{p}_g$, we represent a trajectory by $K+1$ sampled waypoints $\mathcal{P}=\{\bm{p}_0,\bm{p}_1,\ldots,\bm{p}_K\}$, where $\bm{p}_0=\bm{p}_s$ and $\bm{p}_K=\bm{p}_g$. For a query point $\bm{p}$, the Hilbert map predicts the latent occupancy value
\begin{equation}
    \hat{z}(\bm{p}) = \bm{\Phi}(\bm{p})^\top \hat{\bm{\theta}},
\end{equation}

and predictive uncertainty
\begin{equation}
    u(\bm{p}) = \bm{\Phi}(\bm{p})^\top \mathrm{cov}(\hat{\bm{\theta}}) \bm{\Phi}(\bm{p}).
\end{equation}

We compare two planning strategies under the same start-goal query. The greedy deterministic strategy uses only the latent occupancy field, while the uncertainty-aware strategy additionally penalizes uncertain regions. Both strategies can be written as the following unified constrained optimization:
\begin{equation}
\small
\begin{aligned}
\mathcal{P}^{\star}_{s}
= \arg\min_{\bm{p}_{1:K-1}}\quad
& \alpha \sum_{k=0}^{K-1}
\left\|\bm{p}_{k+1}-\bm{p}_{k}\right\|_2 \\
& + \beta \sum_{k=0}^{K} c_s(\bm{p}_k)\\
\mathrm{s.t.}\quad
& \left\|\bm{p}_{k+1}-\bm{p}_{k}\right\|_2 \le d_{\max}, \\
& \hfill k=0,\ldots,K-1, \\
& o(\bm{p}_k) \le o_{\max}, \quad k=0,\ldots,K, \\
& \bm{p}_k \in \Omega, \quad k=1,\ldots,K-1 .
\end{aligned}
\label{eq:path_planning_unified}
\end{equation}
and the strategy-dependent map cost is defined as
\begin{equation}
c_s(\bm{p}) =
\begin{cases}
    \hat{z}(\bm{p}), & \text{str.}=\mathrm{G},\\[2mm]
    \dfrac{\hat{z}(\bm{p})}{\sqrt{u(\bm{p})+\epsilon}}, & \text{str.}=\mathrm{U},
\end{cases}
\label{eq:path_planning_strategy_cost}
\end{equation}
in which $\text{str.}\in\{\mathrm{G},\mathrm{U}\}$ denotes the greedy de
terministic and uncertainty-aware strategies, respectively. The small constant $\epsilon$ avoids division by zero. We use the latent occupancy $\hat{z}(\bm{p})$ rather than the saturated probability $o(\bm{p})$ in the objective because the sigmoid output can have vanishing gradients near 0 or 1. The confidence-normalized term $\hat{z}(\bm{p})/\sqrt{u(\bm{p})+\epsilon}$ plays a Sharpe-ratio-like role \cite{sharpe1998sharpe}: free-space predictions are preferred only when they are also confident. 

\begin{figure*}[t]
    \centering
    \includegraphics[width=0.98\linewidth]{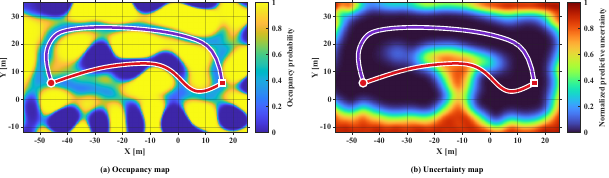}
    \caption{Path planning results on a partial RICH-SLAM map on ColoRadar~6. The left figure shows occupancy probability and the right one shows normalized predictive uncertainty. The red curve denotes the greedy deterministic path, and the purple curve denotes the uncertainty-aware path. Circles and squares denote start and goal positions, respectively. Although the greedy path follows a shorter route through a region that appears traversable in the occupancy map, the uncertainty-aware path avoids the high-uncertainty region and selects a longer but more reliable route, indicating that RICH-SLAM uncertainty can provide useful risk information for downstream planning.}
    \label{fig:uncertainty_aware_path}
\end{figure*}

The optimization is non-convex, so we first compute an A$^*$ path on the queried map as an initial guess and then refine it with continuous trajectory optimization. Figure~\ref{fig:uncertainty_aware_path} shows the resulting paths on both the occupancy map and the uncertainty map. The greedy deterministic path takes a shorter route through a region that appears traversable in occupancy, whereas the uncertainty-aware path chooses a longer route that avoids the high-uncertainty part of the incomplete map. This demonstrates that the predictive uncertainty produced by RICH-SLAM can provide useful information for risk-aware downstream planning.

\subsection{Discussion}
\label{sec:discussion}

\subsubsection{Key Observations}

Overall, the experimental results answer the research questions (\textbf{Q1}--\textbf{Q4}). The experiments lead to several observations beyond the individual quantitative results. First,  compactness is important for the mapping of this study: a moderate number of Hilbert basis functions is sufficient to recover useful continuous occupancy structure, while simply increasing the feature dimension yields diminishing returns. Second, the GP hyperparameters have interpretable effects on the map: the length scale controls spatial smoothness, and the signal variances affect occupancy contrast and confidence. Third, map-aided radar localization is most effective when the environment provides distinctive geometric constraints, and its benefit is more pronounced for heading than for translation. Finally, the same posterior map representation supports both localization and planning with the estimated occupancy and uncertainty.

\subsubsection{Computational Considerations}

The per-iteration cost of RICH-SLAM is dominated by two components: the Kalman filter map update with complexity $\mathcal{O}(MN^2 + M^2N + N^3)$ (Table~\ref{tab:SequentialUpdate}), and the particle filter weighting with complexity $\mathcal{O}(M^2 N)$ per particle. In the experiments, we use $M=256$ basis functions and $N_p=200$ particles. The Kalman filter-based map update is therefore practical for online operation with sparse radar measurements. We note, however, that increasing $M$ to support larger environments or finer spatial resolution will increase the map-update cost, and very large-scale deployment may require a block tiling \cite{kok2018scalable}, or hierarchical map strategy.

\subsubsection{Limitations}

We identify several limitations of this study. First, the current system does not include loop closure, so accumulated drift cannot be globally corrected after revisiting previously mapped areas. This limitation is especially relevant for long trajectories and corridor-like environments. Second, the current implementation assumes a 2D map, which restricts applicability in strongly non-planar scenes; extending the method to 3D would require a larger basis-function set and a higher-dimensional map state. Third, the map domain $\Omega$ is specified in advance, which may be impractical for some exploration tasks where the environment extent is unknown. Adaptive domain selection and submap management are important directions for future work.

\section{Conclusion}
\label{sec:conclusion}

We presented RICH-SLAM, a radar SLAM system that performs simultaneous localization and continuous occupancy mapping from sparse and noisy radar measurements. By integrating Hilbert-space Gaussian process mapping with a Rao-Blackwellized particle filter, the system builds compact continuous maps with posterior uncertainty and uses the map posterior for particle weighting. Experiments on self-collected and public ColoRadar datasets show that \textsc{Hilbert-GP} provides an effective finite-dimensional occupancy representation, that RICH-SLAM can incrementally construct continuous maps from sparse radar beams, and that map-aided particle weighting improves localization especially in geometrically informative environments. The path planning demonstration further illustrates that the resulting occupancy and uncertainty fields can be queried directly as traversability costs.

Several directions for future work remain. First, considering the recent advances in radar sensor technology, extending the framework to 3D could be the next step. Second, incorporating loop closure would improve global consistency over long trajectories and repeated visits. Third, the posterior uncertainty produced by the mapping module opens opportunities for tighter integration with downstream tasks such as active exploration and risk-aware motion planning. Finally, because sonar sensing exhibits characteristics similar to radar, adapting RICH-SLAM to underwater environments~\cite{tardos2002robust} is a promising avenue for marine robotics.

\section*{Appendix}
\label{sec:appendix}

This appendix presents the recursive Laplace approximation as an alternative to the Kalman filter described in Section~\ref{sec:kf_update}.

\subsection*{A.1 Recursive Laplace Approximation}

Following the Laplace framework for Bayesian logistic
models~\cite{bishop2006pattern}, we approximate the log-posterior of
the map parameters $\bm{\theta}$ given measurements
$\tilde{\bm{z}}[i]$ as:
\begin{equation}
\begin{aligned}
    \Psi(\bm{\theta})
    &= \log P(\bm{\theta} \mid \tilde{\bm{z}}[i]) \\
    &= \log P(\tilde{\bm{z}}[i] \mid \bm{\theta})
       + \log P(\bm{\theta})
       + \text{const.}
\end{aligned}
\label{eq:posterior}
\end{equation}
where $P(\tilde{\bm{z}}[i] \mid \bm{\theta})$ is the measurement likelihood and $P(\bm{\theta})$ is the prior. The Laplace approximation seeks the mode $\bm{\theta}_{\mathrm{MAP}}$ of this posterior and fits a Gaussian around it.

The prior is carried forward from the previous time step as a
Gaussian:
\begin{equation}
    \log P(\bm{\theta})
    = -\frac{1}{2}\,
      \Delta\bm{\theta}^{T}\,
      \cov{\hat{\bm{\theta}}[i{-}1]}^{-1}\,
      \Delta\bm{\theta}
      + \text{const.}
\end{equation}
where
$\Delta\bm{\theta}
      = \bm{\theta} - \hat{\bm{\theta}}[i{-}1]$
is the deviation from the prior mean and $\cov{\hat{\bm{\theta}}[i{-}1]}$ is the associated covariance matrix. Because each measurement element $(\tilde{\bm{z}}[i])_n$ is binary with values in $\{-1,+1\}$, we define
\begin{equation}
    y_n = \frac{(\tilde{\bm{z}}[i])_n+1}{2}\in\{0,1\}
\end{equation}
The log-likelihood then factorizes as
\begin{equation}
\begin{aligned}
&\log P(\tilde{\bm{z}}[i] \mid \bm{\theta})
=
\sum_{n=1}^{N}
\Big[
y_n \log \varsigma((\bm{z}[i])_n) \\
&\quad
+
(1-y_n)
\log\!\left(1-\varsigma((\bm{z}[i])_n)\right)
\Big]
\end{aligned}
\end{equation}
where $(\bm{z}[i])_n$ is the latent occupancy logit approximated by $\bm{\Phi}(\bm{p}_n[i])^{T}\bm{\theta}$, consistent with Equation~\eqref{eq:Latent}.

The mode is found iteratively via Newton's method:
\begin{equation}
    \bm{\theta}^{\text{new}}
    = \bm{\theta}
      - \bigl(\nabla\nabla\Psi(\bm{\theta})\bigr)^{-1}
        \nabla\Psi(\bm{\theta})
    \label{eq:newton}
\end{equation}
The required gradient and Hessian of $\Psi(\bm{\theta})$ are given by:
\begin{equation}
    \nabla\Psi(\bm{\theta})
    =
    \sum_{n=1}^{N}
      \left(y_n-\varsigma((\bm{z}[i])_n)\right)
      \bm{\Phi}_n
    -
    \cov{\hat{\bm{\theta}}[i{-}1]}^{-1}
    \Delta\bm{\theta}
    \label{eq:gradient}
\end{equation}
\begin{equation}
\begin{aligned}
    \nabla\nabla\Psi(\bm{\theta})
    =
    -\sum_{n=1}^{N}
      & \varsigma((\bm{z}[i])_n)
        \left(1-\varsigma((\bm{z}[i])_n)\right)
        \bm{\Phi}_n\bm{\Phi}_n^{\top} \\
      & -
        \cov{\hat{\bm{\theta}}[i{-}1]}^{-1}
\end{aligned}
    \label{eq:hessian}
\end{equation}
where $\bm{\Phi}_n$ denotes $\bm{\Phi}(\bm{p}_n[i])$ for brevity. Upon convergence, the posterior is approximated by:
\begin{equation}
    P(\bm{\theta} \mid \tilde{\bm{z}}[i])
    \approx \mathcal{N}\!\Bigl(
      \bm{\theta}_{\mathrm{MAP}},\;
      -\bigl(\nabla\nabla\Psi(
        \bm{\theta}_{\mathrm{MAP}})\bigr)^{-1}
    \Bigr)
    \label{eq:laplace_posterior}
\end{equation}
where $\bm{\theta}_{\mathrm{MAP}}$ is the MAP estimate of $\bm{\theta}$. This Gaussian posterior then serves as the prior at the next time step, enabling recursive map updates as new measurements arrive.

\subsection*{A.2 Complexity Comparison}

Table~\ref{tab:SequentialUpdate} compares the per-step and overall complexity of the two approaches. Since the number of basis functions $M$ typically exceeds the number of measurements $N$ per scan, the dominant cost of the Kalman-based update is $\mathcal{O}(M^2N)$, whereas each Newton step of the Laplace approximation costs $\mathcal{O}(M^3)$. Therefore, for $M>N$ the Kalman update is already cheaper per pass; moreover, it is a closed-form single-pass update, while Newton's method requires several iterations to converge. Both factors make the Kalman-based estimator more suitable for online sequential mapping, and we adopt it throughout RICH-SLAM.

\begin{table}[t]
\centering
\caption{Computational complexity of two sequential map-update methods. $M$ represents the number of basis functions and $N$ is the number of measurements.}
\label{tab:SequentialUpdate}
\scriptsize
\begin{tabular}{lll}
\hline\hline
Method & Recursive Laplace approx.
       & Kalman filter-based est. \\
\hline
\multirow{3}{*}{Steps}
  & Eq.~\eqref{eq:gradient}: $\mathcal{O}(NM + M^2)$
  & Eq.~\eqref{eq:covz}: $\mathcal{O}(M^2 N + MN^2)$ \\
  & Eq.~\eqref{eq:hessian}: $\mathcal{O}(NM^2)$
  & Eq.~\eqref{eq:gain_matrix}: $\mathcal{O}(MN^2 + N^3)$ \\
  & Eq.~\eqref{eq:newton}: $\mathcal{O}(M^3)$
  & Eq.~\eqref{eqUpdateMap}: $\mathcal{O}(MN^2 + M^2 N)$ \\
Overall
  & $\mathcal{O}(NM^2 + M^3)$
  & $\mathcal{O}(MN^2 + M^2 N + N^3)$ \\
\hline\hline
\end{tabular}
\end{table}

\bibliographystyle{IEEEtran}
\bibliography{IEEEexample}

\end{document}